  \RequirePackage{fix-cm}
  \documentclass[twocolumn]{svjour3}          
  \smartqed  
  \usepackage{graphicx}
  \usepackage[authoryear]{natbib}
  \usepackage{tabu,multirow}
  \usepackage[leftcaption]{sidecap}
  \usepackage{amsmath}
  \usepackage{amssymb}
  \usepackage{color}
  \usepackage{enumitem}
  \usepackage{verbatim}
  \usepackage[colorlinks,citecolor=blue]{hyperref}
  
  \usepackage{graphicx}
  \usepackage{subfigure}
  \usepackage{float}  
  \usepackage{textcomp}

  \usepackage{booktabs}
  \usepackage{multirow}
  \usepackage{colortbl}
  \usepackage{xcolor}
  \usepackage{microtype}
  \usepackage{graphicx}
  \usepackage{subfigure}

  \usepackage{eso-pic}
  \usepackage{xspace}
  \makeatletter
  \DeclareRobustCommand\onedot{\futurelet\@let@token\@onedot}
  \def\@onedot{\ifx\@let@token.\else.\null\fi\xspace}

  \makeatother
  
  \usepackage{mathtools}

  \usepackage{multirow}
  \usepackage{array}
  \usepackage{algorithm}
  \usepackage{algorithmic}
  
\newtheorem{assumption}[theorem]{Assumption}
  
\begin{document}
  \sloppy
  
  \title{Rethinking Meta-Learning from a Learning Lens}
  
  
  
  \author{ Jingyao~Wang      \and
           Wenwen~Qiang \and
           Changwen~Zheng \and
           Hui~Xiong\and
           Gang~Hua
  }
  
 \institute{Jingyao Wang, Wenwen Qiang, Changwen Zheng \at
              National Key Laboratory of Space Integrated Information System, Institute of Software Chinese Academy of Sciences, Beijing, China; University of Chinese Academy of Sciences, Beijing, China\\
           \and
          Hui Xiong \at
              Thrust of Artificial Intelligence, Hong Kong University of Science and Technology, Guangzhou, China;
              Department of Computer Science and Engineering, the Hong Kong University of Science and Technology, Hong Kong SAR, China\\
          \and    
          Gang Hua \at
            Amazon.com, Inc., Bellevue, WA, 98004, USA\\
          \and   
          Corresponding author: Wenwen Qiang, \email{qiangwenwen@iscas.ac.cn}
}

\date{Received: date / Accepted: date}

\def\ourconv{RIConv++\xspace}
\def\smallgap{\vspace{0.05in}}
  
\maketitle

\begin{abstract}
Meta-learning seeks to learn a well-generalized model initialization from training tasks to solve unseen tasks. From the ``learning to learn'' perspective, the quality of the initialization is modeled with one-step gradient decent in the inner loop. However, contrary to theoretical expectations, our empirical analysis reveals that this may expose meta-learning to underfitting. To bridge the gap between theoretical understanding and practical implementation, we reconsider meta-learning from the ``Learning'' lens. We propose that the meta-learning model comprises two interrelated components: parameters for model initialization and a meta-layer for task-specific fine-tuning. These components will lead to the risks of overfitting and underfitting depending on tasks, and their solutions—fewer parameters vs. more meta-layer—are often in conflict. To address this, we aim to regulate the task information the model receives without modifying the data or model structure. Our theoretical analysis indicates that models adapted to different tasks can mutually reinforce each other, highlighting the effective information. Based on this insight, we propose TRLearner, a plug-and-play method that leverages task relation to calibrate meta-learning. It first extracts task relation matrices and then applies relation-aware consistency regularization to guide optimization. Extensive theoretical and empirical evaluations demonstrate its effectiveness.

  \end{abstract}

\section{Introduction}
\label{sec:1}

Meta-learning, also known as ``learning to learn'', acquires knowledge from multiple tasks and then adapts to unseen tasks.
Recently, meta-learning has demonstrated tremendous success in various applications, such as affective computing ~\cite{li2023compound}, image classification ~\cite{chen2021metadelta}, and robotics ~\cite{schrum2022mind}.

This work focuses on meta-learning methods based on bi-level optimization \cite{verma2020meta, Metasurvey}. The main approaches include optimization-based \cite{maml, reptile, anil, optimization_rapid} and metric-based methods \cite{protonet, relationnet}. These methods typically aim to learn an effective model initialization, which is subsequently fine-tuned for downstream tasks to produce task-specific models. Let the initialized model be $ \mathcal{F}_{\theta} $, and the task-specific model for the $i$-th task be $ f_\theta^i $. The training data for each task is divided into a support set and a query set (analogous to the training and testing sets in traditional machine learning). Then, the meta-learning process includes two steps: (i) The inner loop, referred to as ``to learn'', aims to derive $ f_\theta^i $ through a single gradient descent step based on $ \mathcal{F}_{\theta} $ and the support set; (ii) The outer loop, referred to as ``learning'', updates $ \mathcal{F}_{\theta} $ based on the performance of $f_\theta^i $ on the query set, again using gradient descent. Notably, the ``one gradient descent step'' reflects the proximity of $ \mathcal{F}_{\theta} $ and $ f_\theta^i $. The bi-level learning process enforces a constraint that the model obtained after one-step gradient descent should perform well on the given task. That is, based on $ \mathcal{F}_{\theta} $, a single gradient descent step should produce the optimal $ f_\theta^i $. Thus, the ``quality'' of the initialized model is primarily modeled as ``one-step gradient descent'', as the ``best'' $ \mathcal{F}_{\theta} $ is the one closest to optimal $ f_\theta^i $.

Both theoretically and methodologically, meta-learning based on bi-level optimization has made significant progress \cite{chen2019closer,maml,anil}. 
However, understanding meta-learning from ``learning a good initialization model'' has a gap with practical implementation.
First, there is a logical paradox between the objective of meta-learning and its actual implementation during the training phase. According to Eq.\ref{eq:inner}, the constraints aim to obtain the optimal task-specific model. Clearly, achieving this goal with only a single gradient descent step is difficult. From the ``good initialization model'' perspective, however, ``one gradient descent step'' is indeed a crucial component. Second, as shown in \cite{flennerhag2021bootstrapped}, distilling the information contained in the parameters of task-specific models obtained after ``multiple gradient descent steps'' into those obtained after ``one gradient descent step'' can significantly enhance meta-learning performance on downstream tasks. 
Our experiments in \textbf{Figure \ref{fig:motivation}} also demonstrate that meta-learning optimization relying on single-step gradient descent is insufficient (See \textbf{Section \ref{sec:rethink_ex}} for more details).
These pieces of evidence show that understanding meta-learning from learning a ``good initialization model'' is inadequate.

\begin{figure}
    \centering
    \subfigure[Modeling of meta-learning model $\mathcal{F}_\theta$]{
        \includegraphics[width=\columnwidth]{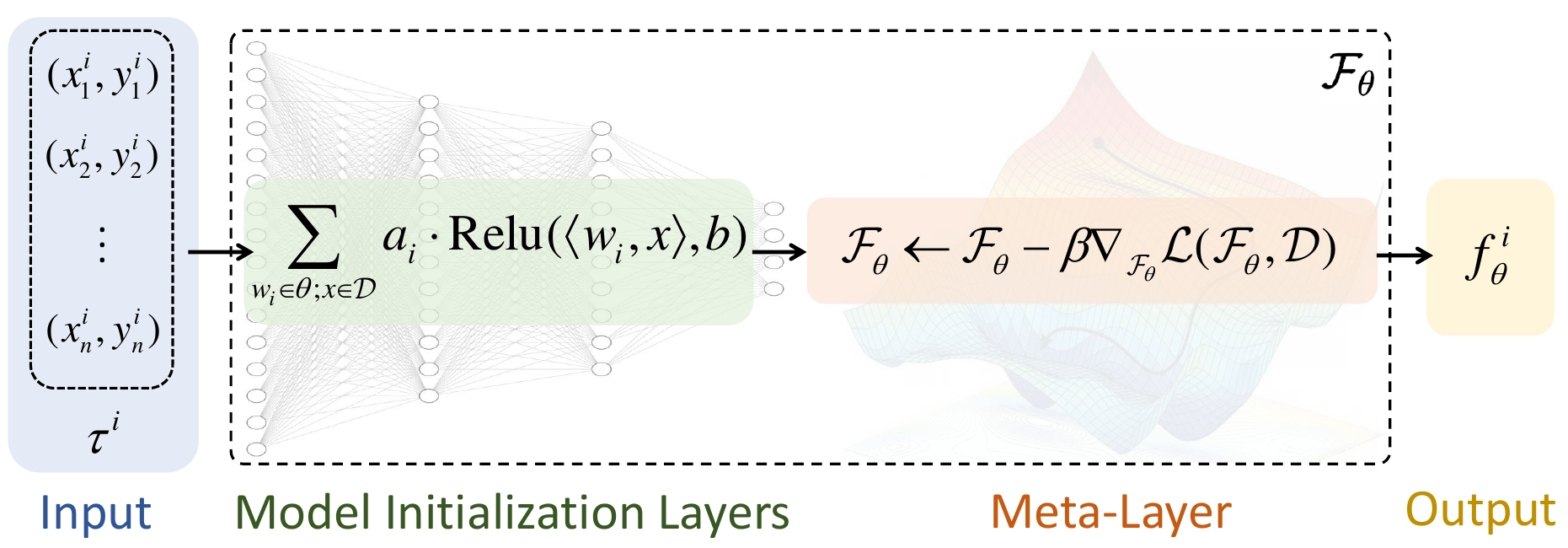}}\label{fig:model_formulation}
        \hfill
    \subfigure[Learning of meta-learning model $\mathcal{F}_\theta$]{
        \includegraphics[width=\columnwidth]{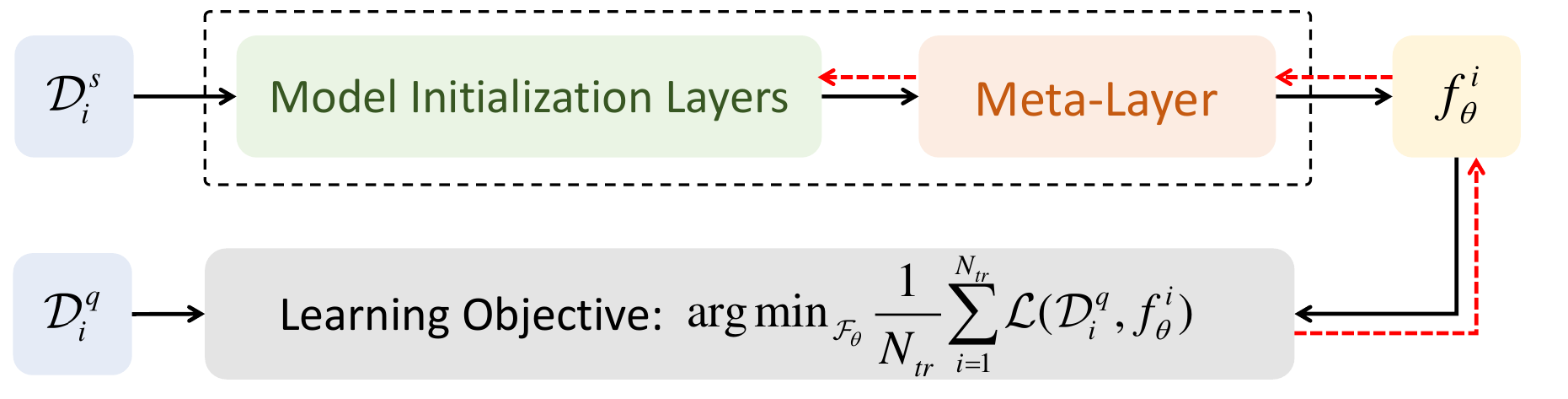}}\label{fig:model_process}
    \caption{Reformulation of meta-learning model $\mathcal{F}_\theta$. (a) briefly shows how to model $\mathcal{F}_\theta$. (b) show the learning process under the modeling of $\mathcal{F}_\theta$ in (a). The black solid line represents the forward computation process, while the red dashed line indicates the backward propagation process.}
    \label{fig:app_o}
\end{figure}

To bridge the above gap, we rethink meta-learning from the ``Learning'' lens to unify the theoretical understanding of meta-learning with its practical implementation (\textbf{Subsection \ref{sec:rethink_formulation}}). Different from the previous understanding that meta-learning refers to ``learning a good initialization'', we focus on the viewpoint that meta-learning can be explained as learning a model $\mathcal{F}_{\theta}$ that given any task $\tau_i$, outputs a task-specific model $f_\theta^i$ that performs well, i.e., $\mathcal{F}_\theta(\tau_i)=f_{\theta}^i$.
\textbf{The central challenge lies in modeling \(\mathcal{F}_{\theta}\), an issue that prior research has largely overlooked but this paper specifically addresses this.} A natural idea to model \(\mathcal{F}_{\theta}\) is to employ a large MLP to construct $\mathcal{F}_{\theta}$, but the required parameters could be prohibitively large. Drawing inspiration from the enhanced representational power of nonlinear functions \cite{haarhoff1970new,schwartz1969nonlinear}, this issue can be solved by incorporating a nonlinear layer. We propose to use gradient optimization to model this nonlinear layer, called ``meta-layer''. Then, we get $\mathcal{F}_\theta-\frac{\partial\mathcal{L} }{\partial \theta}=f_\theta^i$. Compared with the MLP-based modeling, the meta-layer reduces the parameters of $\mathcal{F}_\theta$ while improving the representational capacity. $\mathcal{F}_\theta$ can be regarded as consisting of the model initialization layers and the meta-layer (\textbf{Figure \ref{fig:app_o}}).
This modeling has two advantages. First, it aligns with the bi-level optimization: (i) Inner-loop: output $f_\theta^i$ via $F_\theta-\frac{\partial\mathcal{L} }{\partial \theta}=f_\theta^i$, (ii) Outer-loop: update $\mathcal{F}_\theta$ by evaluating the performance of multiple outputs $f_\theta^i$.
Second, it is more flexible and bridges the gap between theory and implementation. Specifically, for few-shot tasks, a single meta-layer can be used to avoid over-fitting; for complex tasks, more meta-layers can be used to avoid overfitting.

Under our understanding, the key issue is determining how many meta-layers to use for $\mathcal{F}_\theta$. 
Due to task diversity, it is difficult to find a fixed number of meta-layers to suit all tasks. 
This results in meta-learning models facing modeling errors, i.e., errors caused by model selection \cite{mohri2018foundations} that are difficult to eliminate.
To address this, we propose a proxy to balance the modeling errors.
According to \cite{lecun2015deep}, the key to the accurate prediction of models is to fully learn important features of each task.
Following \cite{pearl2009causality}, important features refer to those directly related to labels and shared by samples within the same class. 
Without considering other errors such as data sampling, the presence of modeling errors may directly affect the ability of the model to extract important features, e.g., underfitting or overfitting of the model can lead to biased learning of important features. Thus, the proxy aims to constrain the model to capture important features of different tasks without changing the model structure.
Through theoretical analysis (\textbf{Theorem \ref{theorem:motivation}}), we prove that the classifier for a specific task in meta-learning can leverage features from similar tasks to promote classification. 
Thus, a good meta-learning model $\mathcal{F}_\theta$ should produce similar outputs for task-specific models on similar tasks.
\textbf{This inspires us to develop a method for extracting task relations and integrating them into meta-learning to make the model focus on important task features, achieving the proxy}.

Motivated by the above insight, we finally propose Task Relation Learner (TRLearner), a plug-and-play method that leverages task relations to calibrate meta-learning. It first computes a task relation matrix based on task-specific meta-data. These meta-data are extracted via an adaptive sampler to make it contain discriminative information.  Then, it uses a relation-aware consistency regularization to calibrate meta-learning. The regularization term constrains the meta-learning model to produce similar performance after fine-tuning on similar tasks with the obtained matrix, thereby enhancing the model's focus on important features. Theoretical analyses demonstrate that after the introduction of TRLearner, the meta-learning model achieves smaller excess risk and better generalization performance.

The main contributions can be summarized as: \textbf{(i)} We rethink meta-learning from the ``learning'' lens to bridge the gap between theoretical understanding and practical implementation (\textbf{Section \ref{sec:rethinking}}).
\textbf{(ii)} We propose TRLearner, a plug-and-play method that leverages task relations to calibrate meta-learning (\textbf{Section \ref{sec:4}}).
\textbf{(iii)} Theoretical and empirical evaluations demonstrate the effectiveness of TRLearner (\textbf{Sections \ref{sec:5} and \ref{sec:7}}).

\section{Related Work}
\label{sec:2}

Meta-learning seeks to acquire general knowledge from various tasks and then apply the learned knowledge to new tasks. Typical methods can be divided into two categories: optimization-based \cite{maml,reptile} and metric-based methods~\cite{protonet, relationnet}. They both rely on bi-level optimization to learn general knowledge, resulting in remarkable performance on new tasks. 

Optimization-based meta-learning methods aim to learn optimal initialization parameters that facilitate rapid convergence on new tasks. Classic approaches include MAML \citep{maml}, Reptile \citep{reptile}, and MetaOptNet \citep{lee2019meta}. For instance, MAML trains a model that adapts to diverse tasks by sharing initial parameters and applying multiple gradient updates \citep{abbas2022sharp,jeong2020ood,wang2024towards}. In contrast, Reptile also utilizes shared initialization but adopts an approximate update strategy by iteratively fine-tuning the model to approach optimal parameters. MetaOpt, on the other hand, focuses on selecting effective optimizers and learning rates for rapid task adaptation without directly adjusting model parameters.

Metric-based meta-learning methods, in contrast, learn embedding functions that project instances from various tasks into a feature space where non-parametric classification is feasible. This concept has been examined through several approaches that differ in how embeddings are learned and how similarity or distance metrics are defined. Notable methods in this category include the Siamese Network \citep{koch2015siamese}, Matching Network \citep{vinyals2016matching}, Prototypical Network \citep{protonet}, Relation Network \citep{relationnet}, and Graph Neural Network-based models \citep{hospedales2021meta}. Specifically, the Siamese Network maximizes the similarity between two augmented views of the same instance \citep{chen2021exploring}, and Graph Neural Network approaches explore meta-learning through inference on partially observed graphical models \citep{gao2023survey}. The remaining methods and their variants \citep{vinyals2016matching,wang2024image,zhu2022convolutional,wang2024meta} generally seek to construct a metric space where classification is achieved by computing distances to prototype representations.

Despite its adaptability to various scenarios \cite{sun2023meta,li2018learning,yao2021improving,wang2024meta}, meta-learning still may face over-fitting or under-fitting issues on different tasks. Some works ~\cite{jamal2019task,lee2020meta,yao2021improving} proposed addressing these issues by maintaining network overparameterization while enhancing data or its information content. However, these methods rely on augmentation strategies and sufficient training which highly increases the computational overhead. They focus on changing data but ignore the impact of the more essential ``learning to learn'' strategy of meta-learning, where exists a gap between theoretical understanding and practical implementation. In contrast, we rethink the learning paradigm to explore what causes errors and how to eliminate them.

\section{Problem Formulation and Challenge}
\label{sec:3}

\subsection{Problem Settings}

Meta-learning aims to learn an effective $ \mathcal{F}_\theta = h \circ g $ that can adapt to unseen tasks. Here, $g$ is the feature extractor and $h$ is the classifier. 
A meta-learning dataset is divided into two parts, e.g., meta-training task dataset $\mathcal{D}_{tr}$ and meta-test task dataset $\mathcal{D}_{te}$, these two are all assumed to be sampled from an identical task distribution $p(\mathcal{T})$. Moreover, $\mathcal{D}_{tr}$ and $\mathcal{D}_{te}$ have no class overlap. During training, each batch includes $ N_{tr} $ tasks, denoted as $ \left \{ \tau_i \right \}_{i=1}^{N_{tr}} \in \mathcal{D}_{tr} $. Each task $ \tau_i $ comprises a support set $ \mathcal{D}_i^s = \{ (x_{i,j}^s, y_{i,j}^s) \}_{j=1}^{N_i^s} $ and a query set $ \mathcal{D}_i^q = \{ (x_{i,j}^q, y_{i,j}^q) \}_{j=1}^{N_i^q} $. Here, $ (x_{i,j}^{\cdot}, y_{i,j}^{\cdot}) $ is the sample and corresponding label, $ N_i^{\cdot} $ is the number of samples.

The learning mechanism of meta-learning can be regarded as a bi-level optimization process \cite{maml,protonet}. In the first level, it fine-tunes $\mathcal{F}_\theta$ on task $\tau_i$ with the support set $\mathcal{D}_i^s$, obtaining the task-specific model $f_{\theta}^i$ through one-step gradient descent:
\begin{equation}
\label{eq:inner}
\begin{array}{l}
    f^i_{\theta} \gets \mathcal{F}_\theta -\alpha \nabla_{\mathcal{F}_\theta}\mathcal{L}(\mathcal{D}_i^s,\mathcal{F}_\theta), \\[5pt]
    \text{where} \quad \mathcal{L}(\mathcal{D}_i^s,\mathcal{F}_\theta )=\frac{1}{N_i^s} \sum_{j=1}^{N_i^s}y_{i,j}^s\log {\mathcal{F}_\theta}(x_{i,j}^s),
\end{array}
\end{equation}
where $\alpha$ denotes the learning rate. 
In the second level, meta-learning updates the model $\mathcal{F}_\theta$ using the obtained task-specific models $f_\theta^i$ and the query sets $\mathcal{D}_i^q$ of multiple tasks. The objective can be expressed as:
\begin{equation}
\label{eq:outer}
\begin{array}{l}
    \mathcal{F}_\theta \gets \mathcal{F}_\theta-\beta \nabla_{\mathcal{F}_\theta}\frac{1}{N_{tr}}\sum_{i=1}^{N_{tr}}\mathcal{L}(\mathcal{D}_i^q,f_\theta^i),  \\[5pt]
    \text{where} \quad \mathcal{L}(\mathcal{D}_i^q,f_\theta^i)= \frac{1}{N_i^q} \sum_{j=1}^{N_i^q}y_{i,j}^q\log {f_\theta^i }(x_{i,j}^q),
\end{array}
\end{equation}
where $\beta$ is the learning rate. 
Note that $ f_\theta^i $ is derived by taking the derivative of $\mathcal{F}_\theta$, making $ f_\theta^i $ a function of $\mathcal{F}_\theta$. Consequently, updating $\mathcal{F}_\theta$ as described in Eq.\ref{eq:outer} can be interpreted as computing the second derivative of $\mathcal{F}_\theta$.

Generally, existing methods \cite{maml,verma2020meta} understand the aforementioned bi-level optimization process in meta-learning from the perspective of ``learning a good initialization''. 
Specifically, ``Learning a good initialization'' requires the meta-learning model to adapt quickly to tasks. Achieving this hinges on the effective realization of ``adapt quickly'' which can be modeled as the ``one gradient descent'' in the first level. Therefore, through this understanding, we can obtain that ``one gradient descent'' is the key to the implementation of a ``good model initialization''.

\paragraph{\textbf{Existing Challenge}}
However, there exists a gap between the above understanding and practical implementation. 
Specifically, empirical evidence suggests that existing methods are prone to underfitting \cite{wang2024towards}, which occurs when first-level updates are inadequate, e.g., one-step gradient descent. However, the above formulation views the ``one gradient descent'' as essential for achieving a good initialization, rather than a potential cause of underfitting. 
Second, according to \cite{flennerhag2021bootstrapped}, in scenarios like reinforcement learning, transferring the information from task-specific models learned through future steps, i.e., ``multiple gradient descent steps'', into those obtained from ``a single gradient descent step'' enhances meta-learning performance on downstream tasks. 
Thus, the current understanding of meta-learning remains limited.

\begin{figure*}
\begin{center}
\centerline{\includegraphics[width=\textwidth]{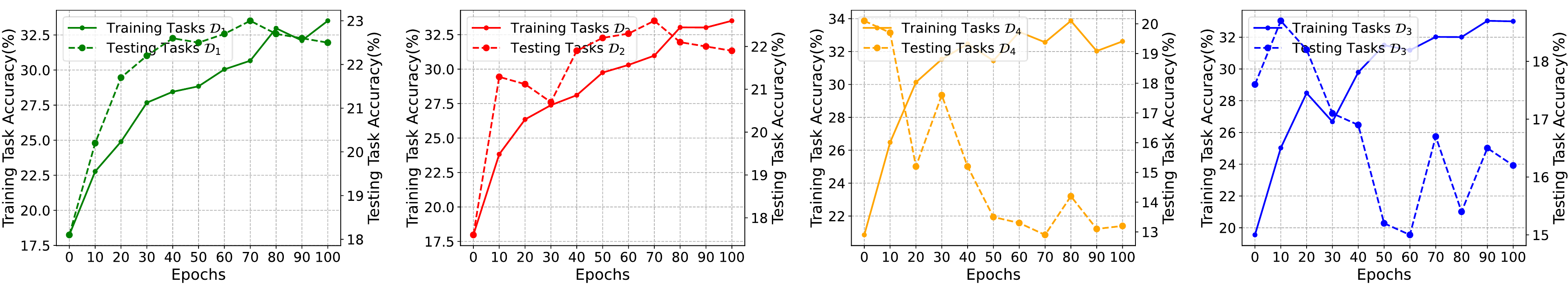}}
\caption{Motivating evidence about the performance of the model on $\mathcal{D}_1$-$\mathcal{D}_4$. Each group of tasks has a different sampling score, i.e., 0.74, 0.68, 0.31, and 0.29 respectively. Higher sampling scores indicate greater task complexity.}
\label{fig:motivation}
\end{center}
\end{figure*}

\subsection{Empirical Evidence.}
\label{sec:rethink_ex}

To verify the limitations of existing methods that understand meta-learning from learning a ``good initialization model'', we conduct a toy experiment. 
It evaluates the performance of meta-learning models relying on a single ``meta-layer'' across different tasks, i.e., whether face overfitting and underfitting according to tasks.

Specifically, we first sample 20 sets of tasks from miniImagenet \cite{miniImagenet} based on \cite{wang2024towards}. 
we first randomly select 20 sets of tasks from the miniImagenet dataset \cite{miniImagenet} following the method in \cite{wang2024towards}. The adaptive sampler \cite{wang2024towards} used here is the same as mentioned in \textbf{Subsection \ref{sec:4.1}} which aims to sample task-specific meta-data for each task. It conducts three metrics, i.e., task diversity, task entropy, and task difficulty, which consider four important indicators to perform task sampling, i.e., intra-class compaction, inter-class separability, feature space enrichment, and causal invariance. In this experiment, we use the first metric to calculate the score of the 20 sets of sampled tasks. Among these, we select the two tasks with the highest sampling scores as $\mathcal{D}_1$-$\mathcal{D}_2$ and the two with the lowest scores as $\mathcal{D}_3$-$\mathcal{D}_4$. 
The higher the sampling scores, the more complex the task. Then, we perform four rounds of data augmentation on $\mathcal{D}_1-\mathcal{D}_2$. Next, we train MAML \cite{maml} on these sets of tasks, i.e., fine-tuning the model with one gradient descent step in the inner loop. We record the training loss and the accuracy on previously unseen test tasks. The results are shown in \textbf{Figure \ref{fig:motivation}}. From the results, we observe that: (i) models trained on $\mathcal{D}_4$ exhibit an inflection point in training loss but perform worse on the test set, indicating overfitting; (ii) models on $\mathcal{D}_1$ show lower performance after convergence and the test performance gradually improves, indicating underfitting. These results demonstrate that existing methods do face the limitations of overfitting and underfitting depending on tasks.

\section{Problem Analysis and Motivation}
\label{sec:rethinking}

To unify the theoretical understanding of meta-learning with its practical implementation, we revisit meta-learning from the ``learning'' lens (\textbf{Subsection \ref{sec:rethink_formulation}}). 
Based on the analyses, we then conduct theoretical analyses (\textbf{Subsection \ref{sec:rethink_th}}) to explore how to eliminate the limitations of existing meta-learning methods.

\subsection{Rethink Meta-Learning from ``Learning'' Lens}
\label{sec:rethink_formulation}

We focus on the view that meta-learning is to learn a well-generalized model $\mathcal{F}_{\theta}$: given any task $\tau_i$, it can output a task-specific model $f_\theta^i$ that performs well, i.e., $\mathcal{F}_\theta(\tau_i)=f_{\theta}^i$. 
The dataset for task $\tau_i$ is denoted as $\mathcal{D}_i$, then the desired $f_\theta^i$ is to achieve $\min\mathbb{E}_{(x,y) \in \mathcal{D}_i} [ \ell(f_{\theta}^i(x),y) ]$. Having the forms of task $\tau_i$ and task-specific model $f_\theta^i$, the central challenge lies in modeling $\mathcal{F}_{\theta}$, a topic not addressed in existing literature, which this paper focuses on.

A natural idea is to employ an MLP to construct $\mathcal{F}_{\theta}$ since MLP is capable of approximating any continuous function using a series of linear layers \cite{pinkus1999approximation,taud2018multilayer}.
However, the required MLP would be extraordinarily large, affecting applicability. 
First, the parameter count is substantial: for complex tasks, the larger parameter count of \( f_\theta^i \) results in an increase in the parameters of \( \mathcal{F}_\theta \). 
Second, the capacity of the MLP is also considerable: MLPs rely on linear layers to fulfill the need for meta-learning that handles various tasks demands a high representational capability.

To address the above limitations, we aim to reduce the network parameters without compromising its representational ability, thus better modeling \(\mathcal{F}_\theta\). According to \cite{haarhoff1970new,schwartz1969nonlinear}, to improve the representational capability of the network, one approach is to use multiple linear layers, while another is to introduce fewer nonlinear layers. 
For instance, to represent the unit circle, we can use either an infinite series of linear equations \( y = wx + b \) or one nonlinear equation \( x^2 + y^2 = 1 \).
Based on this, we propose introducing nonlinear layers to replace the original linear layers, reducing the parameter count of the network while preserving its representational capability.

Specifically, we propose using the gradient optimization function to implement the nonlinear layer, called the ``meta-layer''. 
Compared to the Relu-based function \cite{daubechies2022nonlinear}, the computational complexity of the nonlinear gradient optimization function is higher \cite{watrous1988learning}.
Then, the meta-learning model $\mathcal{F}_\theta$ can be modeled as consisting of the model initialization layers and a meta-layer (\textbf{Figure \ref{fig:app_o}(a)}). 
The model initialization layers can be seen as composed of multiple interconnected neurons, e.g., ResNet50. These neurons work together through weighted inputs \(\langle \omega_i, x \rangle\), bias term \(b\), the activation function \(\text{ReLU}\), and scaling factor \(a_i\), as: 
$\mathcal{F}_\theta: \sum_{\omega_i \in \theta, x \in \mathcal{D}} a_i \cdot \text{ReLU}(\langle \omega_i, x_i \rangle, b)$.
The construction of the meta-layer is motivated by the first-level optimization within meta-learning. 
It is defined by the loss function \(\mathcal{L}(\cdot)\), gradient computation \(\nabla_{\mathcal{F}_\theta}\), and learning rate \(\beta\), as: 
$\mathcal{F}_\theta\gets \mathcal{F}_\theta-\beta \nabla_{\mathcal{F}_\theta}\mathcal{L}(\mathcal{F}_\theta,\mathcal{D})$.
From this implementation, the learnable parameters in the meta-layer are the same as in model initialization layers. Therefore, the learnable parameters of \(\mathcal{F}_\theta\) are those in model initialization layers.
For the learning process (\textbf{Figure \ref{fig:app_o}(b)}),
the dataset contains one meta-training task dataset and one meta-test task dataset. Each task \(\tau_i\) consists of a support set \(\mathcal{D}_i^s\) and a query set \(\mathcal{D}_i^q\). First, the model \(\mathcal{F}_\theta\) takes \(\mathcal{D}_i^s\) as input and outputs task-specific model \(f_\theta^i\), i.e., \(\mathcal{F}_\theta(\tau_i) = f_{\theta}^i\). Then, we evaluate the performance of multiple outputs, i.e., the loss \(\mathcal{L}(\mathcal{D}_i^q, f_\theta^i)\) on each query set \(\mathcal{D}_i^q\), and update \(\mathcal{F}_\theta\).
The objective is:
\begin{equation}\label{eq:obj_metalayer}
    \arg\min_{\mathcal{F}_\theta} \frac{1}{N_{tr}}\sum_{i=1}^{N_{tr}}\mathcal{L}(\mathcal{D}_i^q,f_\theta^i).
\end{equation}
This modeling achieves two advantages.
First, it aligns with the bi-level optimization of meta-learning: (i) The forward computation process (black line in \textbf{Figure \ref{fig:app_o}(b)}) is to obtain the task-specific model, i.e., $F_\theta-\frac{\partial\mathcal{L} }{\partial \theta}=f_\theta^i$ (inner-loop). (ii) The back-propagation process (red line in \textbf{Figure \ref{fig:app_o}(b)}) updates $\mathcal{\theta}$ using multiple $f_\theta^i$ (outer-loop).
Second, it unifies the theoretical understanding with practical implementation, which is more flexible.
Specifically, we can flexibly adjust the number of meta-layers to improve model performance: (i) In few-shot tasks, $\mathcal{F}_\theta$ uses only one meta-layer to balance the parameters and the data volume, avoiding overfitting. (ii) In complex tasks, $\mathcal{F}_\theta$ uses more meta-layers to support more sufficient learning, avoiding underfitting.

\subsection{Theoretical Insights: Method Motivation}
\label{sec:rethink_th}

Based on the above analyses, the key to improving meta-learning is determining the appropriate number of meta-layers. 
The optimal number of meta-layers varies across tasks, making it difficult to define a fixed number suitable for all meta-learning tasks. Consequently, the persistent modeling errors \cite{mohri2018foundations} from meta-layer depth selection adversely affect model performance.
To address this, we conduct theoretical analyses to explore a proxy that can balance the modeling errors.
As noted by \cite{lecun2015deep}, accurate prediction relies on learning the important features of each task. Drawing from \cite{pearl2009causality}, the important features are directly linked to the labels and commonly shared among samples within the same class, e.g., the color of webbed feet or the shape of wings in classifying ``ducks''. 
Ignoring factors such as data sampling errors, the modeling errors directly impact the ability of models to extract important features. For instance, modeling errors may cause underfitting or overfitting, which results in biased learning of important features.
Thus, we aim to constrain
the model to capture important features of different tasks
without changing the model structure to achieve the proxy.
Considering the multi-task joint learning mechanism \cite{maml} of meta-learning, we wonder whether it is possible to capture similar or even identical important features from similar tasks.  
To explore this assertion, we consider a simple scenario of two binary classification tasks $\tau_i$ and $\tau_j$ in the same meta-learning batch, with data variables $X_i$ and $X_j$, and label variables $Y_i$ and $Y_j$ from $\left\{ { \pm 1} \right\}$. Meanwhile, each task in the same batch contains both task-shared and task-specific factors \cite{pearl2009causality}. Then, we have: 
\begin{theorem} \label{theorem:motivation}
Regardless of the correlation between the label variables $Y_i$ and $Y_j$, the classifier for task $\tau_i$ assigns non-zero weights for task-specific factors of task $\tau_j$ with importance $\zeta \propto \text{sim}(X_i,X_j)$ achieve better performance, where $\text{sim}(\cdot)$ is the similarity between $\tau_i$ and $\tau_j$. 
\end{theorem}
\textbf{Theorem \ref{theorem:motivation}} shows that the optimal classifier for a specific task leverages information from other tasks to promote learning. The promotion effect is stronger if the tasks are more similar with the weight $\text{sim}(X_i,X_j)$. Based on this, we propose enforcing task-specific model outputs to be similar on similar tasks, the meta-learning model can obtain optimal classifier and capture effective features. See \textbf{Appendix \ref{sec:app_B}} for detailed proofs and analyses. 
\textbf{This inspires us to leverage task relations to highlight important features, enhancing the performance of $\mathcal{F}_\theta$ across all tasks.}

\begin{figure*}[t]
\begin{center}
\centerline{\includegraphics[width=\textwidth]{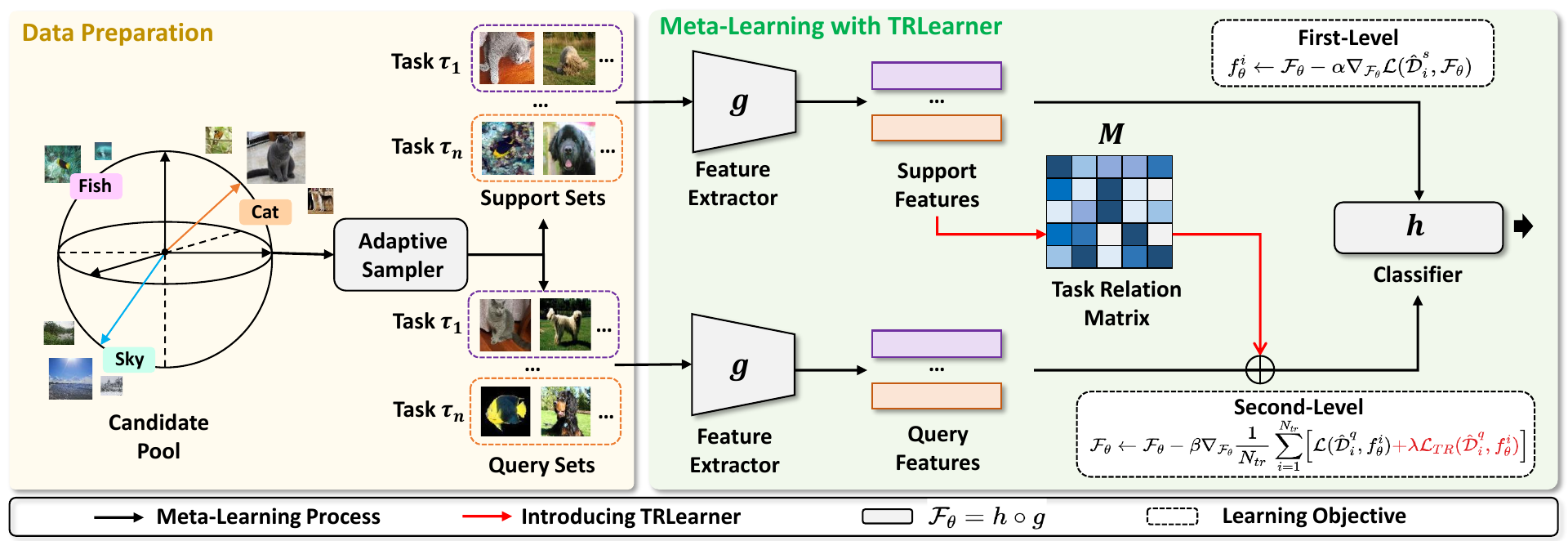}}
\caption{Illustration of meta-learning with TRLearner. TRLearner uses the task relation matrix $\mathcal{M}$ and the regularization term $\mathcal{L}_{TR}$ to calibrate optimization. The black line is for the original meta-learning process, while the red line represents the calibration by TRLearner. The pseudo-code is provided in \textbf{Algorithm \ref{alg:1}}.}
\label{fig:framework}
\end{center}
\end{figure*}

\section{Method}
\label{sec:4}

Based on this insight, we propose Task Relation Learner (TRLearner), which uses task relation to calibrate meta-learning.
Specifically, we first extract the task relation matrix from the sampled task-specific meta-data (\textbf{Subsection \ref{sec:4.1}}). The elements in this matrix reflect the similarity between tasks. Then, we introduce a relation-aware consistency regularization with the obtained matrix to calibrate meta-learning optimization (\textbf{Subsection \ref{sec:4.2}}). Based on \textbf{Theorem \ref{theorem:motivation}}, the regularization term constrains the outputs of meta-learning model $\mathcal{F}_\theta$ on similar tasks achieve similar performance, enforcing the model focus on important features. Finally, in \textbf{Subsection \ref{sec:4.3}}, we introduce the overall objective of meta-learning with TRLearner.
The framework and pseudo-code are shown in \textbf{Figure \ref{fig:framework}} and \textbf{Algorithm \ref{alg:1}}.

\subsection{Extracting Task Relations}
\label{sec:4.1}

We first discuss how to obtain the task relation matrix $\mathcal{M}=\{m_{ij}\}_{i=1,j\ne i}^{N_{tr}}$ between different tasks. 
Each element $m_{ij}$ quantifies the similarity between tasks $\tau_i$ and $\tau_j$. $N_{tr}$ denotes the number of tasks.
Note that directly calculating similarity from all the data within the tasks to obtain $\mathcal{M}$ may cause errors due to sampling randomness and distribution shifts \cite{wang2024towards}. 
Therefore, we propose using a learnable multi-headed similarity layer to acquire $\mathcal{M}$.

Specifically, we first obtain meta-data for each task that reflects the discriminative information using an adaptive sampler \cite{wang2024towards}. 
The higher the sampling scores, the more discriminative the samples are, the greater the sampling probability. We denote the meta-data for task $\tau_i$ as: $\hat{\mathcal{D}}_i^s$ for support set and $\hat{\mathcal{D}}_i^q$ for query set.
Next, we use a multi-headed layer with parameters $\mathcal{W}$ to obtain task relations $\mathcal{M}$. Taking task $\tau_i$ and task $\tau_j$ as examples, we first input the extracted support sets, i.e., $\hat{\mathcal{D}}_i^s$ and $\hat{\mathcal{D}}_j^s$, into the meta-learning model $\mathcal{F}_{\theta}$. Through the feature extractor $g$, we obtain the corresponding task representations $g(\hat{\mathcal{D}}_i^s)$ and $g(\hat{\mathcal{D}}_j^s)$. Then, we calculate their similarity $m_{ij}$ with $\mathcal{W}$:
\begin{equation}\label{eq:relation}
    m_{i,j}=\frac{1}{K}\sum_{k=1}^{K} \cos (\omega_k \odot g(\hat{\mathcal{D}}_i^s), \omega_k \odot g(\hat{\mathcal{D}}_j^s) ),
\end{equation}
where $K$ denotes the number of heads, $\odot$ denotes the Hadamard product, and $\left \{ \omega_k \right \}_{k=1}^K $ denotes the learnable vectors of $\mathcal{W}$ which shares the same dimensions as the task representation, e.g., $g(\hat{\mathcal{D}}_i^s)$. It aims to accentuate variations across the different dimensions within the vector space. Note that the initial weights of the matrix are all 1, i.e., $\omega_k=1$. By calculating the relation between each two tasks in the same batch, we obtain the task relation matrix $\mathcal{M}$.

\begin{algorithm*}
\caption{Meta-Learning with TRLearner}
\label{alg:1}
\textbf{Input}: Task distribution $p(\mathcal{T})$; Randomly initialize meta-learning model $f_{\theta}$ with a feature extractor $g$ and multi-heads $h$; Initialize task relation matrix $\mathcal{M}=\mathbb{I}^{N_{tr} \times N_{tr}} $ \\
\textbf{Parameter}: Number of tasks for one batch $N_{tr}$; Learning rates $\alpha$ and $\beta$ for the learning of $f_{\theta}$; Loss weight $\lambda$ for the relation-aware consistency regularization term \\
\textbf{Output}: Meta-learning model $\mathcal{F}_{\theta}$
\begin{algorithmic}[1] 
\WHILE{not coverage} 
\STATE Sample tasks $\tau \sim \left \{ \tau_i \right \}_{i=1}^{N_{tr}} $ from $p(\mathcal{T})$ via the adaptive task sampler
\hspace*{\fill}$\triangleright$ Task Construction
\FOR{all $\tau_i$}
\STATE Obtain the support set $\mathcal{D}_i^s=\left \{ (x_{i,j}^s,y_{i,j}^s) \right \}_{j=1}^{N_i^s}$ for task $\tau_i$
\STATE Obtain the query set $\mathcal{D}_i^q=\left \{ (x_{i,j}^q,y_{i,j}^q) \right \}_{j=1}^{N_i^q}$ for task $\tau_i$
\STATE Update task relation matrix $\mathcal{M}=\{m_{ij}\}_{i=1,j\ne i}^{N_{tr}}$ via Eq.\ref{eq:relation} \hspace*{\fill}$\triangleright$ Calculate Task Relation
\STATE Update the task-specific model $f^i_{\theta}$ using the support set $\mathcal{D}_i^s$ of task $\tau_i$ via Eq.\ref{eq:inner_new} \hspace*{\fill}$\triangleright$ Inner-Loop Update
\ENDFOR
\STATE Calculate relation-aware consistency score $\mathcal{L}_{TR}(\hat{\mathcal{D}}_i^q,f_{\theta}^i)$ for each task \hspace*{\fill}$\triangleright$ Calibrate Optimization Process
\STATE Update meta-learning model $f_{\theta}$ using all the query sets $\mathcal{D}^q$ in a single batch with $\mathcal{L}_{TR}$ via Eq.\ref{eq:outer_new} \hspace*{\fill}$\triangleright$ Outer-Loop Update
\ENDWHILE
\STATE \textbf{return} solution
\end{algorithmic}
\end{algorithm*}

\subsection{Calibrating Meta-Learning}
\label{sec:4.2}

In this subsection, we illustrate how the relation-aware consistency regularization is designed to enforce the meta-learning model focus on important features.
According to \textbf{Subsection \ref{sec:rethink_th}}, a well-designed meta-learning model \(\mathcal{F}_\theta\) should output similar results on similar tasks. Based on this, we propose a relation-aware consistency regularization term $\mathcal{L}_{TR}$. It constrains the task-specific models on similar tasks to perform similarly based on $\mathcal{M}$, enforcing $\mathcal{F}_\theta$ to learn important features. For task $\tau_i$, it can be expressed as:
\begin{equation}
    \mathcal{L}_{TR}(\hat{\mathcal{D}}_i^q,f_{\theta}^i)=\frac{1}{N_i^q}\sum_{j=1}^{N_i^q}\ell(\frac{\sum_{p=1,p\ne i}^{N_{tr}}m_{ip}f_\theta^p(x_{ij}) }{\sum_{q=1,q\ne i}^{N_{tr}}m_{iq}},y_{i,j} ), 
\end{equation}
where $m_{ip}$ is the strength of the relation between task $\tau_i$ and $\tau_p$. 
$\ell(\cdot)$ is the loss that promotes the alignment of the ground truth with the weighted average prediction obtained from all other task-specific models. 
Thus, $\mathcal{L}_{TR}$ encourages $\mathcal{F}_\theta$ to reinforce the interconnections among task-specific models.
Notably, its effectiveness lies in its ability to leverage task relations to emphasize important features—essentially filtering the task information—and thus maintaining effectiveness even when tasks are highly diverse (\textbf{Appendix \ref{sec:discussion}}).

\subsection{Overall Objective}
\label{sec:4.3}

In this subsection, we present how to embed TRLearner and how it calibrates the optimization process of meta-learning.
 
We begin by explaining how we embed TRLearner into meta-learning, i.e., the network structure. We adopt a multi-head neural network architecture consisting of a feature extractor $g$, a multi-headed similarity layer with weight $\mathcal{W}$, and a classifier $h$. The multi-headed layer is for TRLearner, which is used to extract the task relation matrix $\mathcal{M}$ (Eq.\ref{eq:relation}).

Next, we illustrate how TRLearner calibrates the optimization process of meta-learning. Firstly, we input each support set $\hat{\mathcal{D}}_i^s$ into the model \(\mathcal{F}_\theta\) which outputs task-specific model \(f_\theta^i\) with one meta-layer. Then, we calculate the task relation matrix via Eq.\ref{eq:relation} based on these outputs. Next, we update $\mathcal{F}_\theta$ by evaluating the output $f_\theta^i$ with the relation-aware consistency regularization term $\mathcal{L}_{TR}$ on the each query set $\hat{\mathcal{D}}_i^q$.
For the learning objective, the main difference from Eq.\ref{eq:obj_metalayer} is adding the regularization term $\mathcal{L}_{TR}$ with the matrix $\mathcal{M}$:
\begin{equation}
    \arg\min_{\mathcal{F}_\theta} \frac{1}{N_{tr}}\sum_{i=1}^{N_{tr}}\left [ \mathcal{L}(\hat{\mathcal{D}}_i^q,f_\theta^i)+\lambda \mathcal{L}_{TR}(\hat{\mathcal{D}}_i^q,f_{\theta}^i) \right ],
\end{equation}
where $\hat{\mathcal{D}}_i^{\cdot}$ denotes the meta-data. 
As stated in the fifth paragraph of \textbf{Subsection \ref{sec:rethink_formulation}}, it can be reformulated as a bi-level optimization process.
In the first level, the model $\mathcal{F}_\theta$ follows the same objective as Eq.\ref{eq:inner} but using meta-data. The objective can be expressed as:
\begin{equation}
\label{eq:inner_new}
\begin{array}{l}
    \qquad \qquad f^i_{\theta} \gets \mathcal{F}_\theta -\alpha \nabla_{\mathcal{F}_\theta}\mathcal{L}(\hat{\mathcal{D}}_i^s,\mathcal{F}_\theta), \\[5pt]
    \text{where} \quad \mathcal{L}(\hat{\mathcal{D}}_i^s,\mathcal{F}_\theta )=\frac{1}{N_i^s} \sum_{j=1}^{N_i^s}y_{i,j}^s\log {\mathcal{F}_\theta}(x_{i,j}^s),
\end{array}
\end{equation}
where $\alpha$ denotes the learning rate. 
Obtaining task-specific models, TRLearner calculates the task relation matrix via Eq.\ref{eq:relation}. 
In the second level, we optimize the model $\mathcal{F}_\theta$ with the obtained $\mathcal{M}$ and the regularization term $\mathcal{L}_{TR}$. The objective can be expressed as:
\begin{equation}
\label{eq:outer_new}
\begin{array}{l}
    \scalebox{0.96}{$\mathcal{F}_\theta \gets \mathcal{F}_\theta-\beta \nabla_{\mathcal{F}_\theta}\frac{1}{N_{tr}}\sum_{i=1}^{N_{tr}}\left [ \mathcal{L}(\hat{\mathcal{D}}_i^q,f_\theta^i)+\lambda \mathcal{L}_{TR}(\hat{\mathcal{D}}_i^q,f_{\theta}^i) \right ],$}  \\[5pt]
    \text{where} \quad \mathcal{L}(\hat{\mathcal{D}}_i^q,f_\theta^i)= \frac{1}{N_i^q} \sum_{j=1}^{N_i^q}y_{i,j}^q\log {f_\theta^i }(x_{i,j}^q),
\end{array}
\end{equation}
where $\beta$ is the learning rate and $\lambda$ is the importance weight of $\mathcal{L}_{TR}$. Thus, through the above optimization process, the meta-learning model can utilize additional task relation information to calibrate the optimization process without changing data or model structure.

\section{Theoretical Analysis}
\label{sec:5}

In this section, we conduct theoretical analyses to evaluate the effectiveness of TRLearner. 
We first provide an upper bound on the excess risk, showing that by introducing TRLearner, we can obtain a smaller excess risk (\textbf{Theorem \ref{theorem:1}}). Next, we show that leveraging the accurate task relations achieves better generalization than previous methods that treat all training tasks equally (\textbf{Theorem \ref{theorem:2}}). The related assumptions and proofs are provided in \textbf{Appendix \ref{sec:app_B}}.

First, we provide the maximum limit of excess risk.
\begin{theorem}\label{theorem:1}
    Assume that for every task, the training data $\mathcal{D}_i^{tr}$ contains $N_i^{tr}$ that is approximately greater than or equal to the minimum number of samples found across all tasks, i.e., $N_{sh}$. If the loss function $\ell(\cdot)$ is Lipschitz continuous concerning its first parameter, then for the test task $\tau^{te}$, the excess risk adheres to the following condition:
    \begin{equation}
        \scalebox{0.9}{$\sum_{(x,y)\in\mathcal{D}^{te}}\left [ \ell(\mathcal{F}_\theta^*(x),y)-\ell(\mathcal{F}_\theta(x),y) \right ] \le \sigma +\sqrt{\frac{\mathcal{R}(\mathcal{H})}{N_{sh}N_{tr}\sigma ^k} }$}, 
    \end{equation}
    where $N_{tr}$ denotes the number of tasks, while the other symbols, e.g., $\sigma$, $k$, etc., are the same as in Assumption \ref{assumption:1}.
\end{theorem}
It suggests that incorporating task relations can close the distance between training and test risks, resulting in a decrease in the excess risk as the number of training tasks increases. 

Next, we prove that obtaining an accurate task relation matrix $ \mathcal{M}$ can enhance the OOD generalization of meta-learning. Specifically, we denote the the task relation matrix obtained via TRLearner as $\mathcal{M}$, and the matrix of previous methods as $\check{\mathcal{M}} $, where all elements are set to 1. We get:
\begin{theorem}\label{theorem:2}
   Consider the function class $\mathcal{H}$ that satisfies Assumption \ref{assumption:1} and the same conditions as Theorem \ref{theorem:1}, define $r(\mathcal{F}_{\theta}^*,\mathcal{M})$ as the excess risk with task relation matrix $\mathcal{M}$, we have $\underset{\mathcal{F}_{\theta}^*}{\inf} \underset{h \in \mathcal{H}}{\sup} r(\mathcal{F}_\theta^{*},\mathcal{M})-\underset{\mathcal{F}_\theta^*}{\inf} \underset{h \in \mathcal{H}}{\sup} r(\mathcal{F}_\theta^{*},\check{\mathcal{M}}) <0$.
\end{theorem}
\textbf{Theorem \ref{theorem:2}} shows that introducing $\mathcal{M}$ achieves better generalization compared to $\check{\mathcal{M}}$, i.e., having smaller excess risk. Thus, our TRLearner effectively enhances the generalization performance of meta-learning with theoretical support.

\section{Experiments}
\label{sec:7}

To evaluate the effectiveness of TRLearner, we conduct experiments on  (i) regression (\textbf{Subsection \ref{sec:7.1}}), (ii) image classification (\textbf{Subsection \ref{sec:7.2}}), (iii) drug activity prediction (\textbf{Subsection \ref{sec:7.3}}), and (iv) pose prediction (\textbf{Subsection \ref{sec:7.4}}), and (v) OOD tasks (\textbf{Subsection \ref{sec:app_F_4}}). We introduce the experimental settings and datasets in each corresponding subsection. We also conduct ablation studies and visualization analyses to evaluate how TRLearner works and why it performs well (\textbf{Subsection \ref{sec:7.5}} and \textbf{Subsection \ref{sec:7.6}}). 

We apply TRLearner to multiple meta-learning methods, e.g., MAML \cite{maml}, ProtoNet \cite{protonet}, MetaSGD \cite{Meta-sgd}, ANIL \cite{anil}, and T-NET \cite{tnet}. For comparison, we consider the regularizers which handle meta-learning, i.e., Meta-Aug \cite{metaaug}, MetaMix \cite{yao2021improving}, Dropout-Bins \cite{jiang2022role} and MetaCRL \cite{wang2023hacking}, and the SOTA methods proposed for generalization, i.e., Meta-Trans \cite{bengio2019meta}, MR-MAML \cite{mrmaml}, iMOL \cite{imol}, OOD-MAML \cite{oodmaml}, and RotoGBML \cite{rotogbml}. All results are averaged from five runs on NVIDIA V100 GPUs. More details are provided in \textbf{Appendices \ref{sec:O}-\ref{sec:app_C}}

\section{Implementation and Architecture}
\label{sec:app_E}
Within the meta-learning framework, we utilize the Conv4 architecture \cite{maml} as the basis for the feature extractor. After the convolution and filtering steps, we sequentially apply batch normalization, ReLU activation, and $2 \times 2$ max pooling (achieved via stride convolutions). The final output from the feature extractor's last layer is then fed into a softmax layer with $N_{tr}$ heads as classifiers. For one batch of training, we use different heads to participate in the training of task-specific models and introduce relation-aware consistency regularizers to participate in the update of the second layer. These network architectures undergo a pretraining phase and remain unchanged during the training process. Notably, as described in \cite{jiang2022role}, we employ a different architecture for pose prediction experiments. This model consists of a fixed encoder with three convolutional blocks and an adaptive decoder with four convolutional blocks. Each block includes a convolutional layer, batch normalization, and ReLU activation. For the optimization process, we use the Adam optimizer \citep{kingma2014adam} to train our model, with momentum set at 0.8 and weight decay at $0.7\times10^{-5}$. The initial learning rate for all experiments is 0.1, with the option for linear scaling as needed.

\begin{table}[t]
\begin{center}
\caption{Performance (MSE) comparison on the Sinusoid and Harmonic regression. The best results are highlighted in \textbf{bold}, and TRLearner's results are highlighted in \colorbox{orange!10}{orange}.}
\label{tab:regression}
\begin{small}
\begin{sc}
\resizebox{\linewidth}{!}{
\begin{tabular}{l|c|c|c|c}
\toprule
\textbf{Model} & \textbf{(Sin,5-shot)} & \textbf{(Sin,10-shot)} & \textbf{(Har,5-shot)} & \textbf{(Har,10-shot)} \\
\midrule
MR-MAML   &  0.581 $\pm$ 0.110 & 0.104 $\pm$ 0.029 & 0.590 $\pm$ 0.125 & 0.247 $\pm$ 0.089 \\
Meta-Trans & 0.577 $\pm$ 0.123 & 0.097 $\pm$ 0.024 & 0.576 $\pm$ 0.116 & 0.231 $\pm$ 0.074 \\
iMOL & 0.572 $\pm$ 0.107 & 0.083 $\pm$ 0.018 & 0.563 $\pm$ 0.108 & 0.228 $\pm$ 0.062 \\
OOD-MAML & 0.553 $\pm$ 0.112 & 0.076 $\pm$ 0.021 & 0.552 $\pm$ 0.103 & 0.224 $\pm$ 0.058 \\
RotoGBML & 0.546 $\pm$ 0.104 & 0.061 $\pm$ 0.012 & 0.539 $\pm$ 0.101 & 0.216 $\pm$ 0.043 \\
\midrule
MAML  & 0.593 $\pm$ 0.120 & 0.166 $\pm$ 0.061 & 0.622 $\pm$ 0.132 & 0.256 $\pm$ 0.099 \\
MAML+Meta-Aug & 0.531 $\pm$ 0.118 & 0.103 $\pm$ 0.031 & 0.596 $\pm$ 0.127 & 0.247 $\pm$ 0.094 \\
MAML+MetaMix  & 0.476 $\pm$ 0.109 & 0.085 $\pm$ 0.024 & 0.576 $\pm$ 0.114 & 0.236 $\pm$ 0.097 \\
MAML+Dropout-Bins  & 0.452 $\pm$ 0.081 & 0.062 $\pm$ 0.017 & 0.561 $\pm$ 0.109 & 0.235 $\pm$ 0.056 \\
MAML+MetaCRL  & 0.440 $\pm$ 0.079 & 0.054 $\pm$ 0.018 & 0.548 $\pm$ 0.103 & 0.211 $\pm$ 0.071 \\
\rowcolor{orange!10}\textbf{MAML+TRLearner} & \textbf{0.400 $\pm$ 0.064} & \textbf{0.052 $\pm$ 0.016} & \textbf{0.539 $\pm$ 0.101} & \textbf{0.204 $\pm$ 0.037} \\
\midrule
ANIL & 0.541 $\pm$ 0.118 & 0.103 $\pm$ 0.032 & 0.573 $\pm$ 0.124 & 0.205 $\pm$ 0.072 \\
ANIL+Meta-Aug & 0.536 $\pm$ 0.115 & 0.097 $\pm$ 0.026 & 0.561 $\pm$ 0.119 & 0.197 $\pm$ 0.064 \\
ANIL+MetaMix & 0.514 $\pm$ 0.106 & 0.083 $\pm$ 0.022 & 0.554 $\pm$ 0.113 & 0.184 $\pm$ 0.053 \\
ANIL+Dropout-Bins  & 0.487 $\pm$ 0.110 & 0.088 $\pm$ 0.025 & 0.541 $\pm$ 0.104 & 0.179 $\pm$ 0.035 \\
ANIL+MetaCRL & \textbf{0.468 $\pm$ 0.094} & 0.081 $\pm$ 0.019 & 0.533 $\pm$ 0.083 & 0.153 $\pm$ 0.031 \\
\rowcolor{orange!10}\textbf{ANIL+ TRLearner} & 0.471 $\pm$ 0.081 & \textbf{0.075 $\pm$ 0.023} & \textbf{0.517 $\pm$ 0.074} & \textbf{0.134 $\pm$ 0.028} \\
\midrule
MetaSGD  & 0.577 $\pm$ 0.126 & 0.152 $\pm$ 0.044 & 0.612 $\pm$ 0.138 & 0.248 $\pm$ 0.076 \\
MetaSGD+Meta-Aug & 0.524 $\pm$ 0.122 & 0.138 $\pm$ 0.027 & 0.608 $\pm$ 0.126 & 0.231 $\pm$ 0.069 \\
MetaSGD+MetaMix  & 0.468 $\pm$ 0.118 & 0.072 $\pm$ 0.023 & 0.595 $\pm$ 0.117 & 0.226 $\pm$ 0.062 \\
MetaSGD+Dropout-Bins  & 0.435 $\pm$ 0.089 & 0.040 $\pm$ 0.011 & 0.578 $\pm$ 0.109 & 0.213 $\pm$ 0.057 \\
MetaSGD+MetaCRL & 0.408 $\pm$ 0.071 & 0.038 $\pm$ 0.010 & 0.551 $\pm$ 0.104 & 0.195 $\pm$ 0.042 \\
\rowcolor{orange!10}\textbf{MetaSGD+TRLearner} & \textbf{0.391 $\pm$ 0.057} & \textbf{0.024 $\pm$ 0.008} & \textbf{0.532 $\pm$ 0.101} & \textbf{0.176 $\pm$ 0.027} \\
\midrule
T-NET  & 0.564 $\pm$ 0.128 & 0.111 $\pm$ 0.042 & 0.597 $\pm$ 0.135 & 0.214 $\pm$ 0.078 \\
T-NET+Meta-Aug & 0.521 $\pm$ 0.124 & 0.105 $\pm$ 0.031 & 0.584 $\pm$ 0.122 & 0.207 $\pm$ 0.063 \\
T-NET+MetaMix  & 0.498 $\pm$ 0.113 & 0.094 $\pm$ 0.025 & 0.576 $\pm$ 0.119 & 0.183 $\pm$ 0.054 \\
T-NET+Dropout-Bins  & 0.470 $\pm$ 0.091 & 0.077 $\pm$ 0.028 & 0.559 $\pm$ 0.113 & 0.174 $\pm$ 0.035 \\
T-NET+MetaCRL  & 0.462 $\pm$ 0.078 & 0.071 $\pm$ 0.019 & 0.554 $\pm$ 0.112 & 0.158 $\pm$ 0.024 \\
\rowcolor{orange!10}\textbf{T-NET+TRLearner} & \textbf{0.443 $\pm$ 0.058} & \textbf{0.066 $\pm$ 0.012} & \textbf{0.543 $\pm$ 0.102} & \textbf{0.144 $\pm$ 0.013} \\
\bottomrule
\end{tabular}}
\end{sc}
\end{small}
\end{center}
\end{table}

\begin{table*}[t]
\begin{center}
\caption{Performance (accuracy $\pm 95\%$ confidence interval) of image classification on SFSL settings, i.e., (5-way 1-shot and 5-way 5-shot) miniImagenet and (20-way 1-shot and 20-way 5-shot) Omniglot, and CFSL settings, i.e., miniImagenet $\to$ CUB and miniImagenet $\to$ Places. The best results are highlighted in \textbf{bold}. The ``$\setminus $'' denotes that the result is not reported.}
\label{tab:classification}
\vspace{0.1in}
\label{tab:app_F1_1}
\begin{small}
\begin{sc}
\resizebox{\linewidth}{!}{
\begin{tabular}{l|cc|cc|cc|cc}
\toprule
\multirow{2}{*}{\textbf{Model}} 
     & \multicolumn{2}{c|}{\textbf{Omniglot}} & \multicolumn{2}{c|}{\textbf{miniImagenet}} & \multicolumn{2}{c|}{\textbf{miniImagenet$\to $CUB}} & \multicolumn{2}{c}{\textbf{miniImagenet$\to $Places}} \\ 
     & \textbf{20-way 1-shot} & \textbf{20-way 5-shot} & \textbf{5-way 1-shot} & \textbf{5-way 5-shot} & \textbf{5-way 1-shot} & \textbf{5-way 5-shot} & \textbf{5-way 1-shot} & \textbf{5-way 5-shot} \\
\midrule
Meta-Trans & 87.39 $\pm$ 0.51 & 92.13 $\pm$ 0.19 & 35.19 $\pm$ 1.58 & 54.31 $\pm$ 0.88 & 36.21 $\pm$ 1.36 & 52.78 $\pm$ 1.91 & 31.97 $\pm$ 0.52 & $\setminus $ \\
MR-MAML & 89.28 $\pm$ 0.59 & 95.01 $\pm$ 0.23 & 35.01 $\pm$ 1.60 & 55.06 $\pm$ 0.91 & 35.76 $\pm$ 1.27 & 50.85 $\pm$ 1.65 & 31.23 $\pm$ 0.48 & 46.41 $\pm$ 1.22 \\
iMOL & 92.89 $\pm$ 0.44 & 97.58 $\pm$ 0.34 & 36.27 $\pm$ 1.54 & 57.14 $\pm$ 0.87 & 37.14 $\pm$ 1.17 & 51.21 $\pm$ 1.01 & 32.44 $\pm$ 0.65 & 47.55 $\pm$ 0.94 \\
OOD-MAML & 93.01 $\pm$ 0.50 & 98.06 $\pm$ 0.27 & 37.43 $\pm$ 1.47 & 57.68 $\pm$ 0.85 & 39.62 $\pm$ 1.34 & 52.65 $\pm$ 0.77 & 35.52 $\pm$ 0.69 & $\setminus $ \\
RotoGBML & 92.77 $\pm$ 0.69 & 98.42 $\pm$ 0.31 & 39.32 $\pm$ 1.62 & 58.42 $\pm$ 0.83 & 41.27 $\pm$ 1.24 & $\setminus $ & 31.23 $\pm$ 0.48 & $\setminus $ \\
\midrule
MAML  &  87.15 $\pm$ 0.61 & 93.51 $\pm$ 0.25 &  33.16 $\pm$ 1.70 & 51.95 $\pm$ 0.97 & 33.62 $\pm$ 1.18 & 49.15 $\pm$ 1.32 & 29.84 $\pm$ 0.56 & 43.56 $\pm$ 0.88 \\
MAML + Meta-Aug & 89.77 $\pm$ 0.62 & 94.56 $\pm$ 0.20 & 34.76 $\pm$ 1.52 & 54.12 $\pm$ 0.94 & 34.58 $\pm$ 1.24 & $\setminus $ & 30.57 $\pm$ 0.63 & $\setminus $ \\
MAML + MetaMix  & 91.97 $\pm$ 0.51 & 97.95 $\pm$ 0.17 & 38.97 $\pm$ 1.81 & 58.96 $\pm$ 0.95 & 36.29 $\pm$ 1.37 & $\setminus $ & 31.76 $\pm$ 0.49 & $\setminus $ \\
MAML + Dropout-Bins & 92.89 $\pm$ 0.46 & 98.03 $\pm$ 0.15 & 39.66 $\pm$ 1.74 & 59.32 $\pm$ 0.93 & 37.41 $\pm$ 1.12 & $\setminus $ & 33.69 $\pm$ 0.78 & $\setminus $ \\
MAML + MetaCRL & 93.00 $\pm$ 0.42 & 98.39 $\pm$ 0.18 & 41.55 $\pm$ 1.76 & 60.01 $\pm$ 0.95 & 38.16 $\pm$ 1.27 & $\setminus $ & 35.41 $\pm$ 0.53 & $\setminus $ \\
\rowcolor{orange!10}\textbf{MAML + TRLearner} & \textbf{94.23 $\pm$ 0.56} & \textbf{98.74 $\pm$ 0.24} & \textbf{42.86 $\pm$ 1.83} & \textbf{61.74 $\pm$ 0.96} & \textbf{40.54 $\pm$ 1.26} & \textbf{54.51 $\pm$ 0.66} & \textbf{36.12 $\pm$ 0.64} & \textbf{48.22 $\pm$ 0.95} \\
\midrule
ProtoNet  & 89.15 $\pm$ 0.46 & 94.01 $\pm$ 0.19 & 33.76 $\pm$ 0.95 & 50.28 $\pm$ 1.31 & 34.28 $\pm$ 1.14 & 48.62 $\pm$ 0.99 & 30.43 $\pm$ 0.57 & 43.40 $\pm$ 0.88 \\ 
ProtoNet + Meta-Aug & 90.87 $\pm$ 0.52 & 94.17 $\pm$ 0.25 & 33.95 $\pm$ 0.98 & 50.85 $\pm$ 1.16 & 35.67 $\pm$ 1.31 & $\setminus $ & 31.27 $\pm$ 0.62 & $\setminus $ \\ 
ProtoNet + MetaMix  & 91.08 $\pm$ 0.51 & 94.32 $\pm$ 0.29 & 34.23 $\pm$ 1.55 & 51.77 $\pm$ 0.89 & 37.19 $\pm$ 1.24 & $\setminus $ & 31.85 $\pm$ 0.64 & $\setminus $ \\ 
ProtoNet + Dropout-Bins & 92.13 $\pm$ 0.48 & 94.89 $\pm$ 0.23 & 34.62 $\pm$ 1.54 & 52.13 $\pm$ 0.97 & 37.86 $\pm$ 1.36 & $\setminus $ & 32.59 $\pm$ 0.53 & $\setminus $ \\ 
ProtoNet + MetaCRL  & 93.09 $\pm$ 0.25 & 95.34 $\pm$ 0.18 & 34.97 $\pm$ 1.60 & 53.09 $\pm$ 0.93 & 38.67 $\pm$ 1.25 & $\setminus $ & 33.82 $\pm$ 0.71 & $\setminus $ \\ 
\rowcolor{orange!10}\textbf{ProtoNet + TRLearner} & \textbf{94.56 $\pm$ 0.39} & \textbf{96.76 $\pm$ 0.24} & \textbf{35.45 $\pm$ 1.72} & \textbf{54.62 $\pm$ 0.95} & \textbf{39.41 $\pm$ 1.26} & \textbf{55.13 $\pm$ 1.32} & \textbf{34.54 $\pm$ 0.64} & \textbf{49.00 $\pm$ 0.74} \\
\midrule
ANIL   & 89.17 $\pm$ 0.56 & 95.85 $\pm$ 0.19 & 34.96 $\pm$ 1.71 & 52.59 $\pm$ 0.96 & 35.74 $\pm$ 1.16 & 49.96 $\pm$ 1.55 & 31.64 $\pm$ 0.57 & 44.90 $\pm$ 1.32 \\
ANIL + Meta-Aug & 90.46 $\pm$ 0.47 & 96.31 $\pm$ 0.17 & 35.44 $\pm$ 1.73 & 56.46 $\pm$ 0.95 & 36.32 $\pm$ 1.28 & $\setminus $ & 32.58 $\pm$ 0.64 & $\setminus $ \\
ANIL + MetaMix  & 92.88 $\pm$ 0.51 & 98.36 $\pm$ 0.13 & 37.82 $\pm$ 1.75 & 59.03 $\pm$ 0.93 &  36.89 $\pm$ 1.34 & $\setminus $ & 33.72 $\pm$ 0.61 & $\setminus $ \\
ANIL + Dropout-Bins  & 92.82 $\pm$ 0.49 & 98.42 $\pm$ 0.14 & 38.09 $\pm$ 1.76 & 59.17 $\pm$ 0.94 & 38.24 $\pm$ 1.17 & $\setminus $ & 33.94 $\pm$ 0.66 & $\setminus $ \\
ANIL + MetaCRL  & 92.91 $\pm$ 0.52 & 98.77 $\pm$ 0.15 & 38.55 $\pm$ 1.81 & 59.68 $\pm$ 0.94 & 39.68 $\pm$ 1.32 & $\setminus $ & 34.47 $\pm$ 0.52 & $\setminus $ \\
\rowcolor{orange!10}\textbf{ANIL+ TRLearner} & \textbf{93.24 $\pm$ 0.48} & \textbf{99.28 $\pm$ 0.21} & \textbf{38.73 $\pm$ 1.84} & \textbf{60.42 $\pm$ 0.95} & \textbf{41.96 $\pm$ 1.24} & \textbf{56.22 $\pm$ 1.25} & \textbf{35.68 $\pm$ 0.61} & \textbf{47.30 $\pm$ 1.30} \\
\midrule
MetaSGD  & 87.81 $\pm$ 0.61 & 95.52 $\pm$ 0.18 & 33.97 $\pm$ 1.34 & 52.14 $\pm$ 0.92 & 33.65 $\pm$ 1.13 & 50.00 $\pm$ 0.84 & 29.83 $\pm$ 0.66 & 45.21 $\pm$ 0.79 \\
MetaSGD + Meta-Aug & 88.56 $\pm$ 0.57 & 96.73 $\pm$ 0.14 & 35.76 $\pm$ 0.91 & 58.65 $\pm$ 0.94 & 34.73 $\pm$ 1.32 & $\setminus $ & 31.49 $\pm$ 0.54 & $\setminus $ \\
MetaSGD + MetaMix  & 93.44 $\pm$ 0.45 & 98.24 $\pm$ 0.16 & 40.28 $\pm$ 1.64 & 60.19 $\pm$ 0.96 & 35.26 $\pm$ 1.21 & $\setminus $ & 32.76 $\pm$ 0.59 & $\setminus $ \\
MetaSGD + Dropout-Bins  & 93.93 $\pm$ 0.40 & 98.49 $\pm$ 0.12 & 40.31 $\pm$ 0.96 & 60.73 $\pm$ 0.92 & 37.49 $\pm$ 1.37 & $\setminus $ & 33.21 $\pm$ 0.67 & $\setminus $\\
MetaSGD + MetaCRL  & 94.12 $\pm$ 0.43 & 98.60 $\pm$ 0.15 & 41.22 $\pm$ 1.41 & 60.88 $\pm$ 0.91 & 38.61 $\pm$ 1.25 & $\setminus $ & 35.83 $\pm$ 0.63 & $\setminus $ \\
\rowcolor{orange!10}\textbf{MetaSGD+TRLearner} & \textbf{94.57 $\pm$ 0.49} & \textbf{99.43 $\pm$ 0.22} & \textbf{41.64 $\pm$ 0.94} & \textbf{62.43 $\pm$ 0.96} & \textbf{39.58 $\pm$ 1.13} & \textbf{57.56 $\pm$ 1.12} & \textbf{36.42 $\pm$ 0.54} & \textbf{48.20 $\pm$ 0.69} \\
\midrule
T-NET  & 87.66 $\pm$ 0.59 & 95.67 $\pm$ 0.20 & 33.69 $\pm$ 1.72 & 54.04 $\pm$ 0.99 & 34.82 $\pm$ 1.17 &  $\setminus $ & 28.77 $\pm$ 0.48 &  $\setminus $ \\
T-NET + MetaMix  & 93.16 $\pm$ 0.48 & 98.09 $\pm$ 0.15 & 39.18 $\pm$ 1.73 & 59.13 $\pm$ 0.99 & 35.42 $\pm$ 1.28 & $\setminus $ & 30.54 $\pm$ 0.57 & $\setminus $ \\
T-NET + Dropout-Bins  & 93.54 $\pm$ 0.49 & 98.27 $\pm$ 0.14 & 39.06 $\pm$ 1.72 & 59.25 $\pm$ 0.97 & 37.22 $\pm$ 1.37 & $\setminus $ & 31.28 $\pm$ 0.61 & $\setminus $\\
T-NET + MetaCRL  & 93.81 $\pm$ 0.52 & 98.56 $\pm$ 0.14 & 40.08 $\pm$ 1.74 & 59.40 $\pm$ 0.98 & 37.49 $\pm$ 1.14 & $\setminus $ & 32.37 $\pm$ 0.55 & $\setminus $ \\
\rowcolor{orange!10}\textbf{T-NET+TRLearner} & \textbf{94.33 $\pm$ 0.54} & \textbf{98.84 $\pm$ 0.17} & \textbf{40.31 $\pm$ 1.75} & \textbf{61.26 $\pm$ 0.97} & \textbf{40.64 $\pm$ 1.29} &  $\setminus $ & \textbf{34.76 $\pm$ 0.62} &  $\setminus $ \\
\bottomrule
\hline
\end{tabular}
}
\end{sc}
\end{small}
\end{center}
\end{table*}

\begin{table*}[t]
\begin{center}
\caption{Performance on drug activity prediction. ``Mean", ``Mde.", and ``$> 0.3$" are the mean, the median value of $R^2$, and the number of analyses for $R^2> 0.3$ stands as a reliable indicator in pharmacology. The best results are highlighted in \textbf{bold}.}
\label{tab:drugprediction}
\begin{small}
\begin{sc}
\resizebox{0.95\linewidth}{!}{
\begin{tabular}{l|ccc|ccc|ccc|ccc|ccc}
\toprule
\multirow{2}{*}{\textbf{Model}}
    & \multicolumn{3}{c|}{\textbf{Group 1}} & \multicolumn{3}{c|}{\textbf{Group 2}} & \multicolumn{3}{c|}{\textbf{Group 3}} & \multicolumn{3}{c}{\textbf{Group 4}} & \multicolumn{3}{c}{\textbf{Group 5 (ave)}} \\ 
    & \textbf{Mean} & \textbf{Med.} & \textbf{$>$ 0.3} & \textbf{Mean} & \textbf{Med.} & \textbf{$>$ 0.3} & \textbf{Mean} & \textbf{Med.} & \textbf{$>$ 0.3} & \textbf{Mean} & \textbf{Med.} & \textbf{$>$ 0.3} & \textbf{Mean} & \textbf{Med.} & \textbf{$>$ 0.3} \\
\midrule
MAML  & 0.371 &  0.315 & 52 & 0.321 &  0.254 & 43 & 0.318 & 0.239 & 44 & 0.348 & 0.281 & 47 & 0.341 & 0.260 & 45 \\
MAML+Dropout-Bins  & 0.410 & 0.376 & 60 & 0.355 & 0.257 & 48 & 0.320 & 0.275 & 46 & 0.370 & 0.337 & 56 & \textbf{0.380} & 0.314 & 52 \\
MAML+MetaCRL  & 0.413 & 0.378 & 61 & 0.360 & 0.261 & 50 & 0.334 & 0.282 & 51 & 0.375 & 0.341 & 59 & 0.371 & 0.316 & 56 \\
\rowcolor{orange!10}\textbf{MAML+TRLearner}  & \textbf{0.418} & \textbf{0.380} & \textbf{62} & \textbf{0.366} & \textbf{0.263} & \textbf{52} & \textbf{0.342} & \textbf{0.285} & \textbf{52} & \textbf{0.379} & \textbf{0.339} & \textbf{59} & 0.378 & \textbf{0.319} & \textbf{56} \\
\midrule
ProtoNet  & 0.361 & 0.306 & 51 & 0.319 & 0.269 & 47 & 0.309 & 0.264 & 44 & 0.339 & 0.289 & 47 & 0.332 & 0.282 & 47 \\
ProtoNet + Dropout-Bins  & 0.391 & 0.358 & 59 & 0.336 & 0.271 & 48 & 0.314 & 0.268 & 45 & 0.376 & 0.341 & 57 & 0.354 & 0.309 & 52 \\
ProtoNet + MetaCRL  & 0.409 & 0.398 & 62 & 0.379 & 0.292 & 52 & 0.331 & 0.300 & 52 & 0.385 & 0.356 & 59 & 0.381 & 0.336 & 56 \\
\rowcolor{orange!10}\textbf{ProtoNet + TRLearner} & \textbf{0.436} & \textbf{0.402} & \textbf{63} & \textbf{0.384} & \textbf{0.306} & \textbf{54} & \textbf{0.357} & \textbf{0.313} & \textbf{53} & \textbf{0.398} & \textbf{0.372} & \textbf{61} & \textbf{0.393} & \textbf{0.348} & \textbf{57} \\
\midrule
ANIL  & 0.355 & 0.296 & 50 & 0.318 & 0.297 & 49 & 0.304 & 0.247 & 46 & 0.338 & 0.301 & 50 & 0.330 & 0.284 & 48 \\
ANIL+MetaMix  & 0.347 & 0.292 & 49 & 0.302 & 0.258 & 45 & 0.301 & 0.282 & 47 & 0.348 & 0.303 & 51 & 0.327 & 0.284 & 48 \\
ANIL+Dropout-Bins  & 0.394 & 0.321 & 53 & 0.338 & 0.271 & 48 & 0.312 & 0.284 & 46 & 0.368 & 0.297 & 50 & 0.350 & 0.271 & 49 \\
ANIL+MetaCRL  & 0.401 & 0.339 & \textbf{57} & 0.341 & \textbf{0.277} & \textbf{49} & 0.312 & 0.291 & \textbf{48} & 0.371 & 0.305 & \textbf{53} & 0.356 & 0.303 & \textbf{51} \\
\rowcolor{orange!10}\textbf{ANIL+TRLearner}  & \textbf{0.402} & \textbf{0.341} & \textbf{57} & \textbf{0.347} & 0.276 & \textbf{49} & \textbf{0.320} & \textbf{0.296} & \textbf{48} & \textbf{0.374} & \textbf{0.306} & \textbf{53} & \textbf{0.364} & \textbf{0.304} & \textbf{51} \\
\midrule
MetaSGD  & 0.389 & 0.305 & 50 & 0.324 & 0.239 & 46 & 0.298 & 0.235 & 41 & 0.353 & 0.317 & 52 & 0.341 & 0.274 & 47 \\
MetaSGD + MetaMix  & 0.364 & 0.296 & 49 & 0.312 & 0.267 & 48 & 0.271 & 0.230 & 45 & 0.338 & 0.319 & 51 & 0.321 & 0.278 & 48 \\
MetaSGD + Dropout-Bins  & 0.390 & 0.302 & 57 & 0.358 & 0.339 & 56 & 0.316 & 0.269 & 43 & 0.360 & 0.311 & 50 & 0.356 & 0.315 & 51 \\
MetaSGD + MetaCRL  & 0.398 & 0.295 & 59 & 0.356 & 0.340 & 59 & 0.321 & 0.271 & 44 & 0.373 & 0.324 & 55 & 0.362 & 0.307 & 54 \\
\rowcolor{orange!10}\textbf{MetaSGD + TRLearner} & \textbf{0.403} & \textbf{0.314} & \textbf{61} & \textbf{0.367} & \textbf{0.351} & \textbf{60} & \textbf{0.345} & \textbf{0.284} & \textbf{46} & \textbf{0.385} & \textbf{0.328} & \textbf{56} & \textbf{0.374} & \textbf{0.319} & \textbf{55} \\
\bottomrule
\end{tabular}}
\end{sc}
\end{small}
\end{center}
\end{table*}

\begin{table}[t]
\begin{center}
\caption{Performance (MSE $\pm $ 95\% confidence interval) comparison on pose prediction, including the 10-shot and 15-shot results. The best results are highlighted in \textbf{bold}.}
\label{tab:poseprediction}
\begin{small}
\begin{sc}
\resizebox{0.9\linewidth}{!}{
\begin{tabular}{l|c|c}
\toprule
\textbf{Model} & \textbf{10-shot} & \textbf{15-shot}\\
\midrule
Meta-Trans & 2.671 $\pm$ 0.248 & 2.560 $\pm$ 0.196 \\
MR-MAML  & 2.907 $\pm$ 0.255 & 2.276 $\pm$ 0.169 \\
\midrule
MAML  & 3.113 $\pm$ 0.241  & 2.496 $\pm$ 0.182 \\
MAML + MetaMix  & 2.429 $\pm$ 0.198 & 1.987 $\pm$ 0.151 \\
MAML + Dropout-Bins  & 2.396 $\pm$ 0.209 & 1.961 $\pm$ 0.134 \\
MAML + MetaCRL & 2.355 $\pm$ 0.200 & 1.931 $\pm$ 0.134 \\
\rowcolor{orange!10}\textbf{MAML + TRLearner} & \textbf{2.334 $\pm$ 0.216} & \textbf{1.875 $\pm$ 0.132} \\
\midrule
ProtoNet  & 3.571 $\pm$ 0.215 & 2.650 $\pm$ 0.210 \\
ProtoNet + MetaMix  & 3.088 $\pm$ 0.204 & 2.339 $\pm$ 0.197 \\
ProtoNet + Dropout-Bins  & 2.761 $\pm$ 0.198 & 2.011 $\pm$ 0.188 \\
ProtoNet + MetaCRL  & 2.356 $\pm$ 0.171 & 1.879 $\pm$ 0.200 \\
\rowcolor{orange!10}\textbf{ProtoNet + TRLearner} & \textbf{2.341 $\pm$ 0.150} & \textbf{1.860 $\pm$ 0.354} \\
\midrule
ANIL  & 6.921 $\pm$ 0.415 & 6.602 $\pm$ 0.385 \\
ANIL + MetaMix   & 6.394 $\pm$ 0.385 & 6.097 $\pm$ 0.311 \\
ANIL + Dropout-Bins   & 6.289 $\pm$ 0.416 & 6.064 $\pm$ 0.397 \\
ANIL + MetaCRL & 6.287 $\pm$ 0.401 & 6.055 $\pm$ 0.339 \\
\rowcolor{orange!10}\textbf{ANIL + TRLearner} & \textbf{6.287 $\pm$ 0.268} & \textbf{6.047 $\pm$ 0.315} \\
\midrule
MetaSGD  & 2.811 $\pm$ 0.239 & 2.017 $\pm$ 0.182 \\
MetaSGD + MetaMix  & 2.388 $\pm$ 0.204 & 1.952 $\pm$ 0.134 \\
MetaSGD + Dropout-Bins  & 2.369 $\pm$ 0.217  & 1.927 $\pm$ 0.120 \\
MetaSGD + MetaCRL  & 2.362 $\pm$ 0.196 & 1.920 $\pm$ 0.191 \\
\rowcolor{orange!10}\textbf{MetaSGD + TRLearner} & \textbf{2.357 $\pm$ 0.188} & \textbf{1.893 $\pm$ 0.176} \\
\midrule
T-NET & 2.841 $\pm$ 0.177 & 2.712 $\pm$ 0.225 \\
T-NET + MetaMix  & 2.562 $\pm$ 0.280 & 2.410 $\pm$ 0.192 \\
T-NET + Dropout-Bins  & 2.487 $\pm$ 0.212 & 2.402 $\pm$ 0.178 \\
T-NET + MetaCRL  & 2.481 $\pm$ 0.274 & 2.400 $\pm$ 0.171 \\
\rowcolor{orange!10}\textbf{T-NET + TRLearner} & \textbf{2.476 $\pm$ 0.248} & \textbf{2.398 $\pm$ 0.167} \\
\bottomrule
\hline
\end{tabular}}
\end{sc}
\end{small}
\end{center}
\end{table}

\begin{table}[t]
\begin{center}
\caption{Evaluation (accuracy $\pm $ 95\% confidence interval) of OOD generalization on Meta-Dataset. The overall results are not the average of ID (in-domain) and OOD (out-of-domain) results, but rather obtained by training on all ten datasets of Meta-Dataset.}
\vspace{0.1in}
\label{tab:app_ood}
\begin{small}
\begin{sc}
\resizebox{\linewidth}{!}{
\begin{tabular}{l|c|c|c}
\toprule
\textbf{Model} & \textbf{Overall} & \textbf{ID} & \textbf{OOD}\\
\midrule
MAML  & 24.51 $\pm$ 0.13 & 31.37 $\pm$ 0.09 & 19.19 $\pm$ 0.10 \\
MAML + MetaMix  & 24.94 $\pm$ 0.15 & 33.91 $\pm$ 0.12 & 20.00 $\pm$ 0.11 \\
MAML + MetaCRL & 29.65 $\pm$ 0.22 & 36.56 $\pm$ 0.15 & 24.71 $\pm$ 0.14 \\
\rowcolor{orange!10}\textbf{MAML + TRLearner} & \textbf{33.01 $\pm$ 0.27} & \textbf{41.12 $\pm$ 0.15} & \textbf{29.49 $\pm$ 0.12} \\
\midrule
ProtoNet & 37.92 $\pm$ 0.19 & 42.18 $\pm$ 0.17 & 30.89 $\pm$ 0.11 \\
ProtoNet + MetaMix & 37.54 $\pm$ 0.21 & 42.56 $\pm$ 0.16 & 31.15 $\pm$ 0.13 \\
ProtoNet + MetaCRL & 38.91 $\pm$ 0.20 & \textbf{44.27 $\pm$ 0.14} & 33.02 $\pm$ 0.12 \\
\rowcolor{orange!10}\textbf{ProtoNet + TRLearner} & \textbf{40.41 $\pm$ 0.21} & 44.18 $\pm$ 0.16 & \textbf{35.15 $\pm$ 0.12} \\
\bottomrule
\hline
\end{tabular}}
\end{sc}
\end{small}
\end{center}
\end{table}

\subsection{Performance on Regression}
\label{sec:7.1}

\paragraph{\textbf{Experimental Setup.}} 
We calculate the Mean Square Error (MSE) on two regression datasets: Sinusoid dataset \cite{jiang2022role} and Harmonic dataset \cite{wang2024towards}. The datasets here consist of data points generated by a variety of sinusoidal functions, with a minimal number of data points per class or pattern. Each data point comprises an input value $ x $ and its corresponding target output value $ y $. Typically, the input values for these data points fluctuate within a confined range, such as between 0 and $ 2\pi $.
In our experiment, we enhance the complexity of the originally straightforward problem by incorporating noise. Specifically, for Sinusoid regression, we adhere to the configuration proposed by \cite{jiang2022role,wang2023hacking}, where the data for each task is formulated as $ A\sin(\omega \cdot x) + b + \epsilon $, with $ A $ ranging from 0.1 to 5.0, $ \omega $ from 0.5 to 2.0, and $ b $ from 0 to $ 2\pi $. Subsequently, we introduce Gaussian observational noise with a mean of 0 and a variance of 0.3 for each data point derived from the target task. Similarly, the Harmonic dataset \citep{lacoste2018uncertainty} is a synthetic dataset sampled from the sum of two sine waves with different phases, amplitudes, and a frequency ratio of 2: $f(x) = a_1 \sin(\omega x + b_1) + a_2 \sin(2\omega x + b_2)$, where $y \sim \mathcal{N}(f(x), \sigma_y^2)$. Each task in the Harmonic dataset is sampled with $\omega \sim \mathcal{U}(5, 7)$, $(b_1, b_2) \sim \mathcal{U}(0, 2\pi)^2$, and $(a_1, a_2) \sim \mathcal{N}(0, 1)^2$. This process finalizes the construction of the dataset for this scenario.

\paragraph{\textbf{Results.}} The results are provided in \textbf{Table \ref{tab:regression}}. From the results, we can observe that (i) TRLearner achieves better results than the SOTA baselines, with average MSE reduced by 0.028 and 0.021. For example, one of the best SOTA variants under the MAML framework is MAML+MetaCRL, which records an MSE of 0.440 on the Sinusoid task. When TRLearner is incorporated (i.e., MAML+TRLearner), the MSE further drops to 0.400. This 0.040 reduction in MSE is representative of the trend across tasks. (ii) TRLearner also shows significant improvements in all meta-learning baselines, with MSE reduced by more than 0.1. For example, looking at the base MAML model without any auxiliary modules, the original MAML reports an MSE of 0.593 in the Sinusoid 5-shot task. After integrating TRLearner, the model (MAML+TRLearner) achieves an MSE of 0.400. These results demonstrate the superiority of TRLearner.

\subsection{Performance on Image Classification}
\label{sec:7.2}

\paragraph{\textbf{Experimental Setup.}} 

We select four benchmark datasets with two experimental settings, including standard few-shot learning (SFSL) and cross-domain few-shot learning (CFSL). For SFSL, we evaluate the average accuracy on two benchmark datasets, including (i) miniImagenet \cite{miniImagenet}, which consists of 100 classes with 50,000/10,000 training/testing images, split into 64/16/20 classes for meta-training/validation/testing and (ii) Omniglot \citep{Omniglot}, which contains 1,623 characters from 50 different alphabets. For CFSL, we train the models on miniImagenet and test the trained models on two different datasets, including (i) CUB \cite{cub}, which encompasses a collection of 11,788 photographs, categorized into 200 distinct bird species, with 5,794 for testing.; (ii) Places \cite{places}, which boasts an extensive library of over 2.5 million images, meticulously categorized into 205 unique scene categories.
Among them, the setting of CFSL strengthens the distribution difference of data during model training and testing, which can better reflect the OOD generalization performance. 
The evaluation metric is the average accuracy.

\paragraph{\textbf{Results.}} 
From the SFSL and CFSL results in \textbf{Table \ref{tab:classification}}, we can observe that: (i) In SFSL, TRLearner achieves stable performance improvement and surpasses other comparison baselines. For example, our method improves by nearly 7\% on MAML and ProtoNet compared to the meta-learning model, and by an average of 2\% compared to the SOTA plug-and-play model without the need for additional networks. (ii) In CFSL, TRLearner always surpasses the SOTA baseline, indicating that it can achieve better generalization improvement without introducing task-specific or label space augmentations required by the baseline. Combined with the trade-off experiment (accuracy vs. training cost) in \textbf{Subsection \ref{sec:vis_trade-off}}, TRLearner achieves the best generalization improvement under the condition of lower computational cost. This further proves the superiority of TRLearner.

\subsection{Performance on Drug Activity Prediction}
\label{sec:7.3}

\paragraph{\textbf{Experimental Setup.}} We assess TRLearner for drug activity prediction using the pQSAR dataset~\cite{martin2019all}, which forecasts compound activity on proteins with 4,276 tasks. Following \cite{martin2019all,yao2021improving}, we divide the tasks into four groups but conduct the ``Group 5'' that contains tasks from the other four groups for average evaluation. 
In line with the method proposed in \cite{martin2019all}, we partition the dataset by placing the training compounds in the support set and the testing compounds in the query set, with task distributions of 4100 for meta-training, 76 for meta-validation, and 100 for meta-testing.
The evaluation metric is the squared Pearson correlation coefficient ($R^2$), indicating the correlation between predictions and ground-truth. We report the mean and median $R^2$ and the count of $R^2$ exceeding 0.3.

\paragraph{\textbf{Results.}} As shown in \textbf{Table \ref{tab:drugprediction}}, TRLearner achieves comparable or better performance to the SOTA baselines across all the groups of data. Considering that drug activity prediction is a more complex task \cite{martin2019all}, TRLearner not only narrows the gap between the $R^2$ Mean and $R^2$ Median scores but also achieves an improvement in the reliability index $R^2 > 0.3$. These results further demonstrate the superior performance of our method in complex scenarios.

\subsection{Performance on Pose Prediction} 
\label{sec:7.4}

\paragraph{\textbf{Experimental Setup.}} We use the Pascal 3D dataset~\cite{xiang2014beyond} as benchmark dataset for pose prediction. The Pascal 3D dataset consists of outdoor images featuring 12 classes of rigid objects selected from the PASCAL VOC 2012 dataset, with each instance annotated with pose attributes such as azimuth, elevation, and camera distance. In addition, the dataset includes pose-annotated images for these same 12 categories sourced from ImageNet. For the pose prediction task, we preprocess the dataset to form 50 categories for meta-training and 15 for meta-testing. Each category comprises 100 grayscale images with a resolution of \(128 \times 128\) pixels. 
The evaluation metric is MSE.

\paragraph{\textbf{Results.}} 

As shown in \textbf{Table \ref{tab:poseprediction}}, introducing TRLearner achieves results comparable to or even exceeding SOTA baselines without additional augmentation, further confirming its effectiveness. In particular, research in pose prediction shows that employing augmentation can expand the dataset and enhance performance \cite{yao2021improving}. The fact that TRLearner delivers similar improvements suggests that leveraging task complementarity enables the model to capture previously overlooked knowledge, thereby boosting its overall performance.

\subsection{OOD Generalization Performance Comparison}
\label{sec:app_F_4}

\paragraph{\textbf{Experimental Setup.}} To demonstrate the effect of TRLearner on improving generalization ability, we strengthened the distribution difference between training and testing tasks to evaluate its improvement on OOD generalization of meta-learning. 
Specifically, in addition to the classification experiments in the cross-domain few-shot learning scenario, we also select a set of benchmark datasets that are most commonly used for OOD generalization verification, i.e., Meta-dataset \cite{triantafillou2019meta}. This benchmark serves as a substantial resource for few-shot learning, encompassing a total of 10 datasets that span a variety of distinct domains. It is crafted to reflect a more authentic scenario by not confining few-shot tasks to a rigid set of ways and shots. The dataset encompasses 10 varied domains, with the initial 8 in-domain (ID) datasets designated for meta-training, which include ILSVRC, Omniglot, Aircraft, Birds, Textures, Quick Draw, Fungi, and VGG Flower. The final 2 datasets are earmarked for assessing out-of-domain (OOD) performance, namely Traffic Signs and MSCOCO. We assess the efficacy of meta-learning models across these 10 domains, utilizing diverse samplers across the entire suite of 10 datasets. We first sample the metadata of training and testing tasks based on the adaptive sampler \cite{wang2024towards}. Then, we record the performance changes of the meta-learning model before and after the introduction of TRLearner. 

\paragraph{\textbf{Results.}} From the results in Table \ref{tab:app_ood}, we can observe that after the introduction of TRLearner, meta-learning achieves a significant performance improvement, reaching 4\% on average. This further illustrates the effect of task relation on OOD generalization.

\subsection{Ablation Study}
\label{sec:7.5}
In this subsection, we provide the results of the ablation studies, including the effect of $\mathbf{\mathcal{L}_{TR}}$, parameter sensitivity, and effect of different meta-layer.


\begin{figure}[t]
    \centering
        \includegraphics[width=0.75\linewidth]{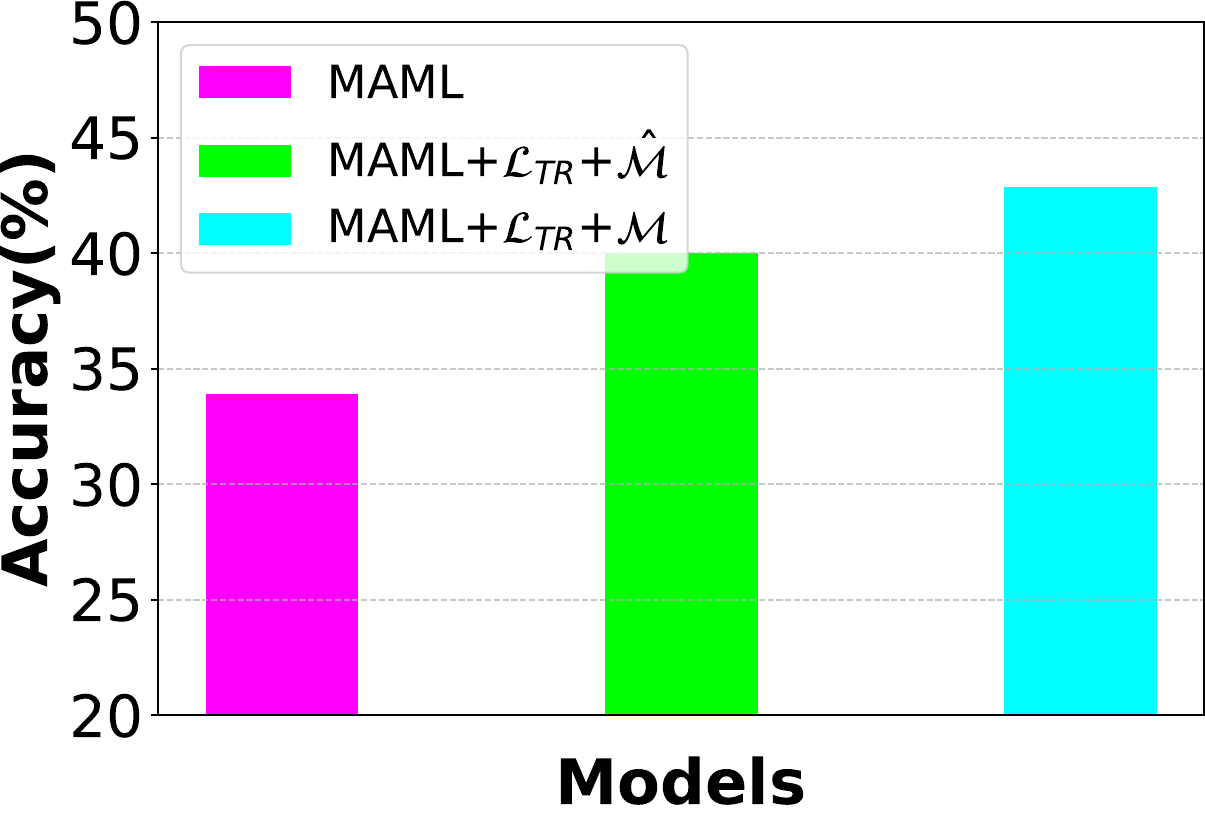}
        \caption{Effect of regularization $\mathcal{L}_{TR}$ on miniImagenet.}
        \label{fig:abla_effect}
\end{figure}

\begin{figure}[t]
    \centering
        \includegraphics[width=0.75\linewidth]{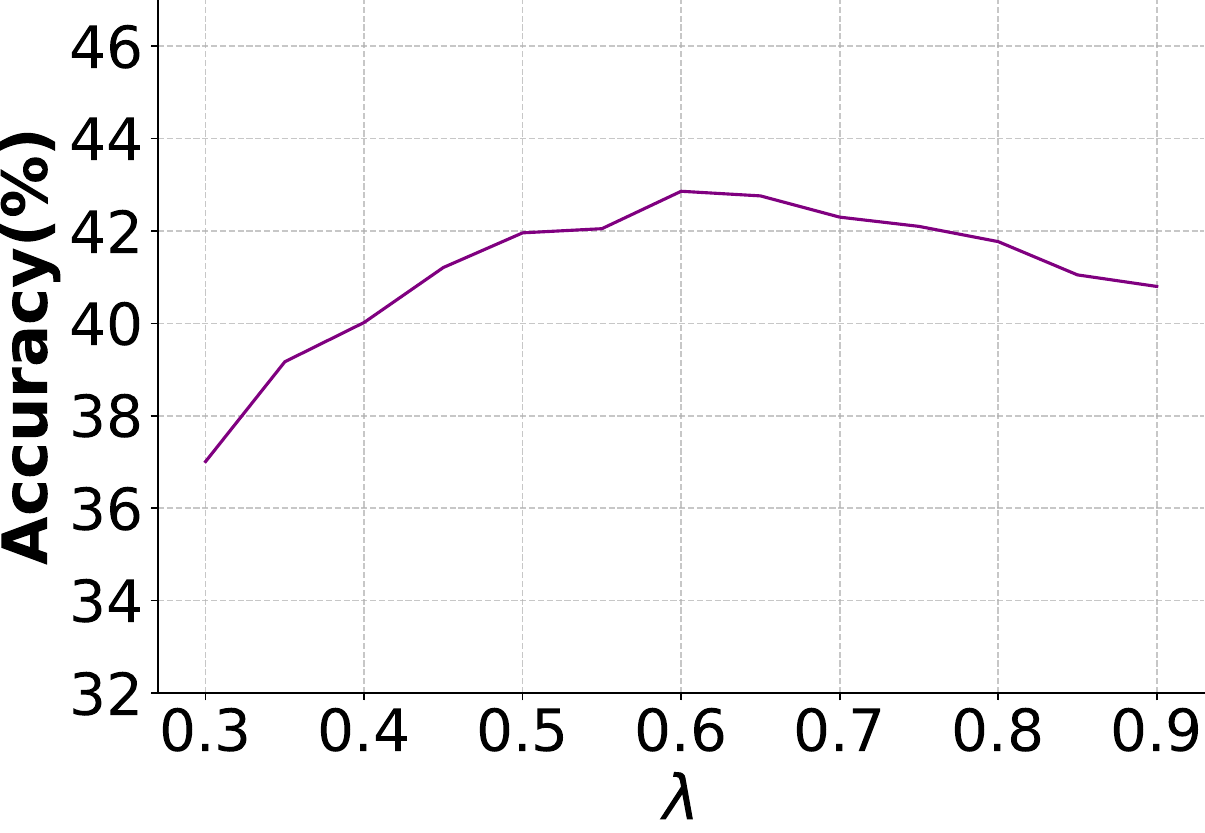}
        \caption{Parameter sensitivity on miniImagenet.}
        \label{fig:abla_para}
\end{figure}

\begin{figure}[t]
\begin{center}
\centerline{\includegraphics[width=0.9\linewidth]{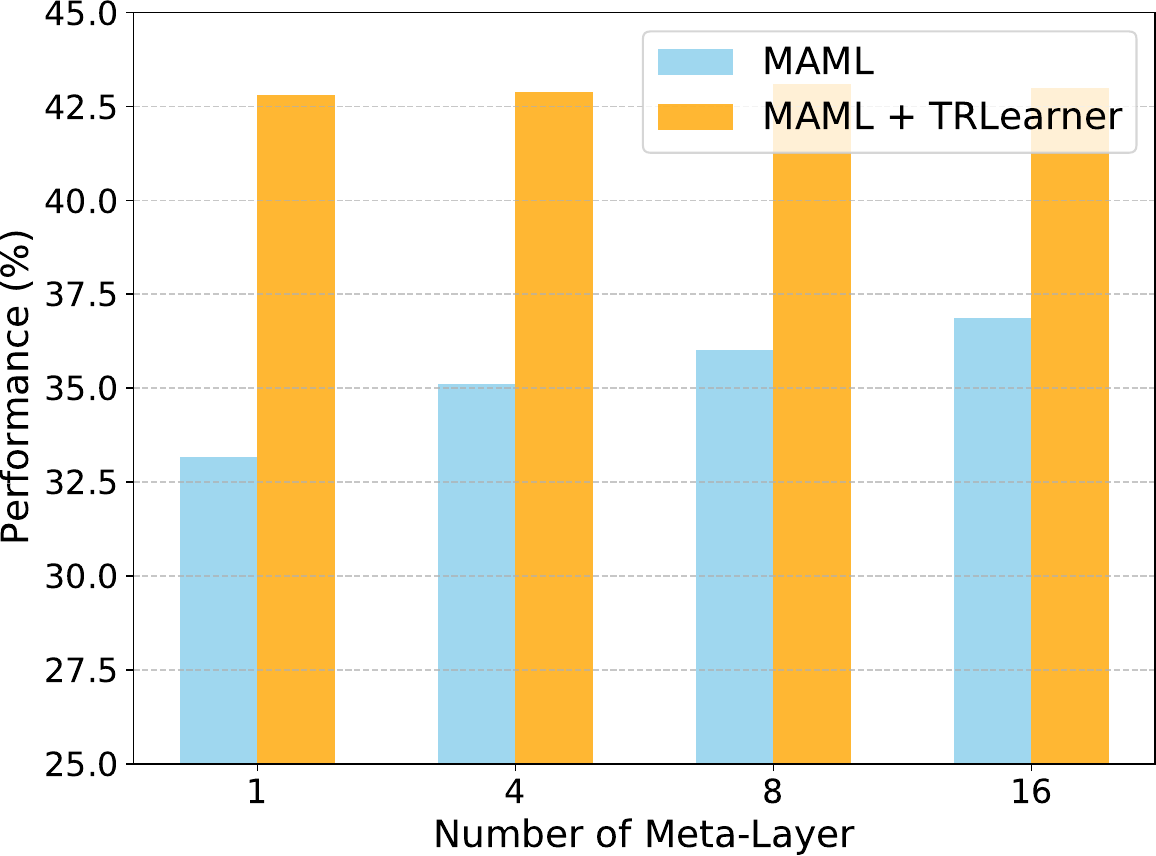}}
\caption{Performance of meta-learning model under the different number of meta-layers. The bars represent the performance of MAML with different meta-layers, i.e., 1, 4, 8, and 16.}
\label{fig:app_motivation_metalayer}
\end{center}
\end{figure}

\subsubsection{Effect of $\mathbf{\mathcal{L}_{TR}}$}
We evaluate the performance of MAML before and after introducing $\mathcal{L}_{TR}$ on miniImagenet, where $\mathcal{L}_{TR}$ is the core of TRLearner. We also evaluate the adaptive learning method of the task relation matrix $\mathcal{M}$ by replacing it with a fixed calculation, i.e., directly calculating the similarity between the sampled meta-data ($\hat{\mathcal{M}}$). The results are shown in \textbf{Figure \ref{fig:abla_effect}}. From the results, we can observe that (i) the model with $\mathcal{L}_{TR}$ has significant improvement and negligible computational overhead; (ii) the adaptively learned $\mathcal{M}$ is more accurate than the fixed calculation. These results prove the effectiveness of the proposed TRLearner.

\subsubsection{Parameter Sensitivity}
We determine the hyperparameters $\lambda$ of the regularization term $\mathcal{L}_{TR}$ by evaluating the impact of different values of $\lambda$ on the performance of MAML+TRLearner with the range $\left [ 0.3, 0.8 \right ] $. The results in \textbf{Figure \ref{fig:abla_para}} show that (i) $\lambda=0.6$ is the best (also our setting), and (ii) TRLearner has minimal variation in accuracy, indicating that hyperparameter tuning is easy in practice.

\subsubsection{Performance Under Different Meta-Layer}
\label{sec:app_F_6}

TRLearner enhances important feature learning by leveraging task relationships, improving the performance of meta-learning models. As described in Subsection \ref{sec:rethink_th}, its design aims to identify a proxy that enables accurate decisions even under modeling errors. Previous experiments have demonstrated TRLearner's performance improvements. To further verify its ability to mitigate modeling errors, we design a set of experiments in this subsection to evaluate the performance of meta-learning models using TRLearner under different meta-layer configurations.
Specifically, we adopt the same experimental setup as in \textbf{Subsection \ref{sec:7.2}}, evaluating on miniImagenet. MAML is selected as the baseline algorithm, with the meta-layer depth set to 1, 4, 8, and 16, respectively, and trained on miniImagenet. Notably, these four configurations share identical training and testing data, differing only in model architecture. According to \cite{mohri2018foundations}, the optimal structure for a specific task varies depending on the task. Models with optimal structures can fully learn task features to support accurate predictions. However, as shown in the experimental results in \textbf{Figure \ref{fig:app_motivation_metalayer}}, the introduction of TRLearner eliminates significant differences in MAML's performance across the four meta-layer settings. This indicates that TRLearner mitigates modeling errors arising from meta-layer depth selection, further demonstrating its effectiveness.

\begin{figure*}
\begin{center}
\centerline{\includegraphics[width=0.9\textwidth]{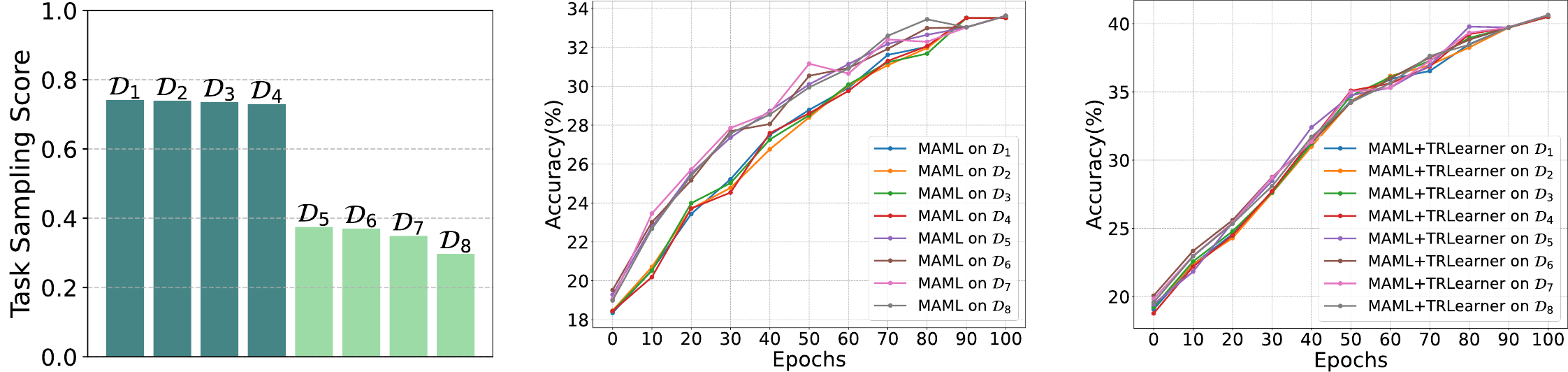}}
\caption{Performance comparison of the motivating experiment after introducing TRLearner. \textbf{Left:} The score of the sampled tasks. \textbf{Middle:} Results of motivating experiment with MAML. \textbf{Right:} Results of motivating experiment with MAML+TRLearner.}
\label{fig:app_motivation}
\end{center}
\end{figure*}

\begin{figure}
\begin{center}
\centerline{\includegraphics[width=0.9\columnwidth]{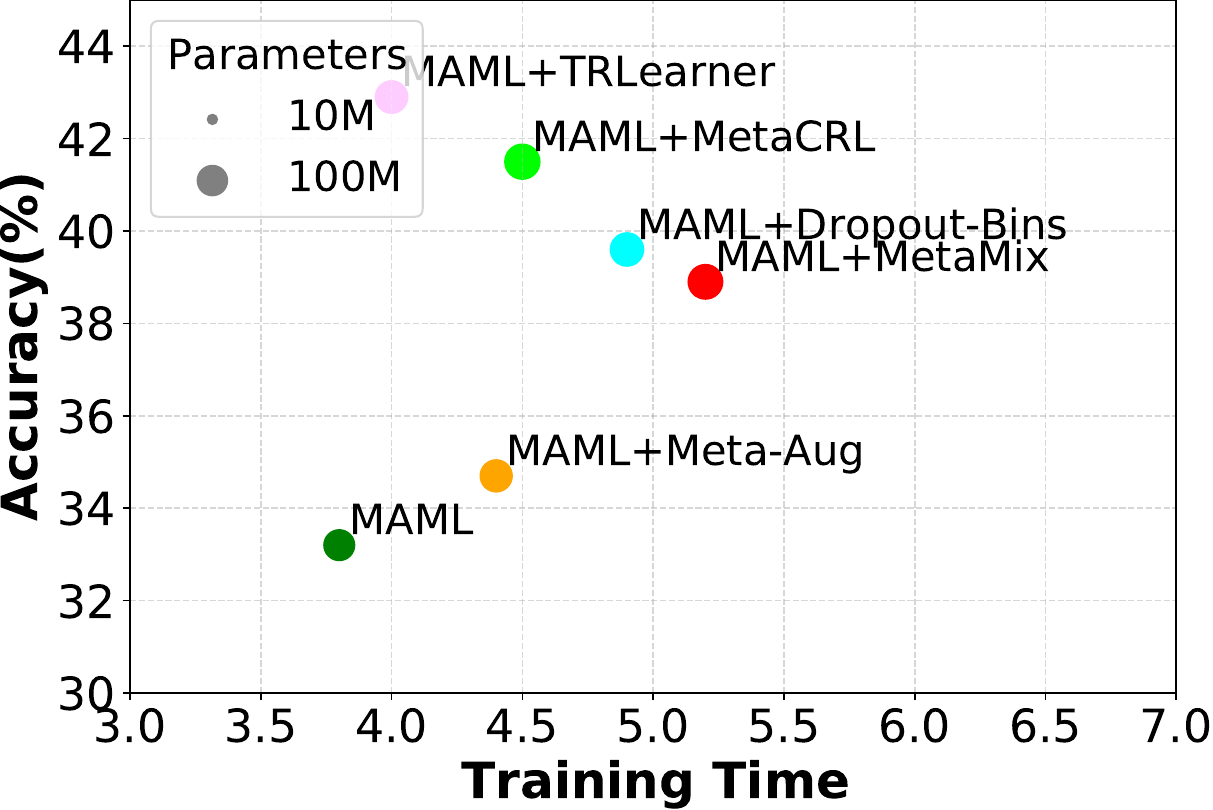}}
\caption{Trade-off performance comparison on miniImagenet. We select MAML as the meta-learning baseline. }
\label{fig:ex_app_trade}
\end{center}
\end{figure}

\begin{figure*}
\begin{center}
\centerline{\includegraphics[width=0.8\textwidth]{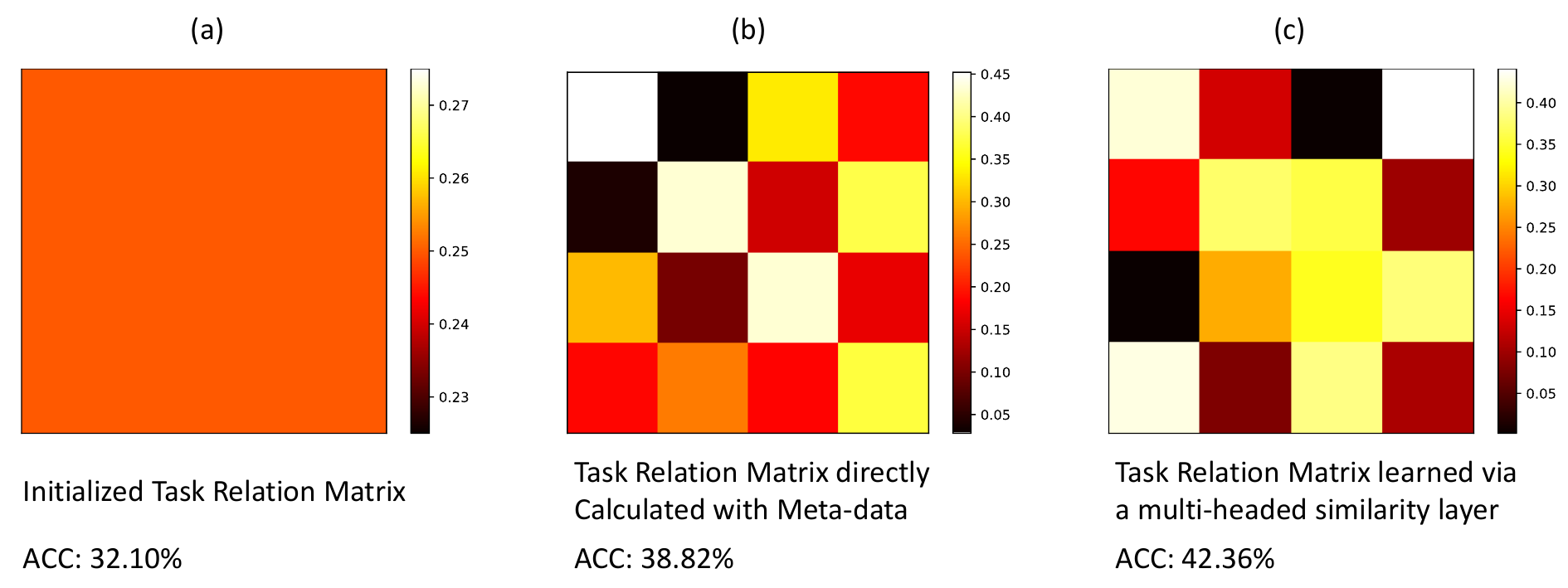}}
\caption{Task Relation Visualization. (a), (b), and (c) respectively represent the initialized task relation matrix, the task relation matrix directly calculated based on the extracted task-specific meta-data, and the task relation matrix further learned using a multi-headed similarity layer. Note that when we visualize the task relation matrix, we normalize the values in each matrix, i.e., the sum of similarity weights between the same task and other tasks is 1.}
\label{fig:app_visualization}
\end{center}
\end{figure*}

\subsection{Visualization Analyses}
\label{sec:7.6}
In this subsection, we provide the results of the visualization analyses to analyze how TRLearner performs well, including trade-off performance, motivating experiments with TRLearner, and task relation visualization. 

\subsubsection{Trade-off Performance}
\label{sec:vis_trade-off}
According to the above analysis, TRLearner improves the generalization of meta-learning in multiple scenarios. Considering that TRLearner may bring additional computational overhead due to the introduction of regularization terms, we evaluate the trade-off performance after introducing TRLearner to ensure its performance in practical applications. Specifically, we use MAML as a baseline, conduct experiments on the miniImagenet dataset, and evaluate its accuracy, training time, and parameter size after introducing different methods. \textbf{Figure \ref{fig:ex_app_trade}} shows the trade-off performance. From the results, we can observe that after introducing TRLearner, the model achieves a significant performance improvement with acceptable calculational cost and parameter size compared to the original framework. Compared with the other baselines, it even achieves faster convergence on the basis of the effect advantage.

\subsubsection{Motivating Results with TRLearner}
\label{sec:vis_motivation}
Considering the randomness in the model training process, we further sample eight sets of data and evaluate the model’s training performance before and after introducing TRLearner.
Specifically, we use the metrics in \cite{wang2024towards} to calculate the score of the 40 sets of sampled tasks. We identify the top four tasks with the highest scores as $\mathcal{D}_1$ to $\mathcal{D}_4$, and the bottom four tasks as $\mathcal{D}_5$ to $\mathcal{D}_8$. Higher sampling scores indicate more complex tasks, providing the model with more information. We then apply four-fold data augmentation to $\mathcal{D}_1$ to $\mathcal{D}_4$. Finally, we assess the MAML model's adaptation on these eight task sets by performing a single gradient descent and recording the accuracy. 
Ideally, $\mathcal{D}_1$ to $\mathcal{D}_4$ not only contain more information, but also further enhance the sample diversity through augmentation. Therefore, the model performs better after training on these four groups of tasks, and there is no overfitting. However, as shown in \textbf{Figure \ref{fig:app_motivation}} middle, in the initial stage of the model, that is, under the constraint of limited training time, the model will be lower than the effect of training on $\mathcal{D}_5$ to $\mathcal{D}_8$. Therefore, it will face the limitation of underfitting since it only performs one step of gradient optimization. This further verifies our point of view, i.e., MAML has the limitations of overfitting and underfitting which is caused by its own learning paradigm. Further, in order to evaluate the impact of introducing TRLearner on the model, we experimented with MAML+TRLearner under the same setting. The results are shown in Figure \ref{fig:app_motivation}. The results show that after the introduction of TRLearner, the overfitting and underfitting phenomena of the model are greatly alleviated.

\subsubsection{Task Relation Visualization}
\label{sec:app_F_5}

In this subsection, we visualize the task relation extracted by TRLearner. Specifically, we visualize the initialized task relation matrix, the task relation matrix directly calculated based on the extracted task-specific meta-data, and the task relation matrix further learned using a multi-headed similarity layer. Taking miniImagenet as an example, we set a training batch including 4 tasks and visualize the matrix and model effect after 100 epochs of training. The visualization results are shown in \textbf{Figure \ref{fig:app_visualization}}. We can observe that the task relation matrix learned based on the multi-headed similarity layer is more accurate, and the meta-learning model learned based on it has the best effect. The results demonstrate the effect of TRLearner and the importance of task relations.

\section{Conclusion}
In this paper, we rethink meta-learning from the ``learning'' lens to unify the theoretical understanding and practical implementation. Through empirical and theoretical analyses, we find that (i) existing meta-learning relying on one meta-layer faces the risks of overfitting and underfitting according to tasks; and (ii) the models adapted to different tasks promote each other where the promotion is related to task relations. Based on these results, we propose TRLearner, a plug-and-play method that uses task relation to calibrate meta-learning optimization. 
Extensive theoretical and empirical analyses demonstrate its effectiveness.


\section*{Data Availability}

The benchmark datasets can be downloaded from the literature cited in each subsection of Section \ref{sec:7}.

\section*{Conflict of interest}

The authors declare no conflict of interest.




%




\bibliographystyle{spbasic}      
\bibliography{reference.bib} 

\clearpage
	
\clearpage
\appendix

\clearpage
\newpage
\appendix

\section*{Appendix}
The appendix provides supplementary information and additional details that support the primary discoveries and methodologies proposed in this paper. It is organized into several sections: 
\begin{itemize}
    \item Appendix \ref{sec:app_B} contains the proofs of the presented theorems.
    \item Appendix \ref{sec:O} provides the details and further analysis about the ``learning'' lens of meta-learning.
    \item Appendix \ref{sec:discussion} provide more discussion about the effectiveness of TRLearner, e.g., with highly diverse tasks.
    \item Appendix \ref{sec:app_C} provides details for all datasets used in the experiments.
    \item Appendix \ref{sec:app_D} provides details for the baselines used in the experiments.
\end{itemize}
Note that before we illustrate the details and analysis, we provide a brief summary of all the experiments conducted in this paper, as shown in Table \ref{tab:app_experiments_summary}.

\begin{table*}[t]
    \centering
    \caption{Illustration of the experiments conducted in this work. All experimental results are obtained after five rounds of experiments.}
    \label{tab:app_experiments_summary}
    \begin{tabular}{p{0.4\textwidth}|p{0.3\textwidth}|p{0.2\textwidth}}
    \toprule
        \textbf{Experiments} & \textbf{Location} & \textbf{Results}\\
    \midrule    
        Motivating experiments & Section \ref{sec:3} and Section \ref{sec:vis_motivation} & Figure \ref{fig:motivation} and Figure \ref{fig:app_motivation}\\
    \midrule    
        Performance on regression problems with two benchmark datasets & Section \ref{sec:7.1} & Table \ref{tab:regression}\\
    \midrule    
        Performance on image classification with two settings, i.e., standard few-shot learning (miniImagenet and Omniglot) and cross-domain few-shot learning (miniImagenet $\to$ CUB and Places) & Section \ref{sec:7.2} & Table \ref{tab:classification} and Table \ref{tab:app_F1_1}\\
    \midrule    
        Performance on drug activity prediction (pQSAR) & Section \ref{sec:7.3} & Table \ref{tab:drugprediction}\\
    \midrule    
        Experiment on pose prediction (Pascal 3D) & Section \ref{sec:7.4} & Table \ref{tab:poseprediction} \\
    \midrule
        Ablation Study-Effect of $\mathcal{L}_{TR}$ & Section \ref{sec:7.5} & Figure \ref{fig:abla_effect}\\
    \midrule
        Ablation Study-Parameter Sensitivity & Section \ref{sec:7.5} & Figure \ref{fig:abla_para}\\
    \midrule
        Trade-off Performance Comparison & Appendix \ref{sec:vis_trade-off} & Figure \ref{fig:ex_app_trade}\\
    \midrule
        OOD Generalization Performance Comparison & Appendix \ref{sec:app_F_4} & Table \ref{tab:app_ood}\\
    \midrule
        Task Relation Visualization & Appendix \ref{sec:app_F_5} & Figure \ref{fig:app_visualization}\\
    \midrule
        Performance under different meta-layer & Appendix \ref{sec:app_F_6} & Figure \ref{fig:app_motivation_metalayer}\\
    \bottomrule
    \end{tabular}
\end{table*}

\section{Proofs}
\label{sec:app_B}
In this section, we provide proofs and analyses of theorems in the main text. Before detailed proofs, we first provide the assumptions to facilitate analysis. Next, we provide proofs of Theorems \ref{theorem:motivation}, \ref{theorem:1}, and \ref{theorem:2}.

\subsection{Assumptions and Discussion}

We first provide the assumptions to facilitate analysis.
\begin{assumption}\label{assumption:1}
    For each task $\tau_i$, the representation $Z_i$ of task $\tau_i$ is derived from the task-specific meta-data $\hat{\mathcal{D}}_i$ via the feature extractor $g$ of meta-learning model $f_{\theta}=h \circ g$, where $h=(h_1, \cdots, h_{N_{tr}})$. Then, we assume:
    \begin{itemize}
        \item $Z_i$ is assumed to be uniformly distributed on $\left [ 0,1 \right ]^k $. 
        \item There exists a universal constant $C$ such that for all $i, j \in N_{tr}$, we have $\left \| h_{i} - h_{j} \right \|_{\infty }  \le  C \cdot  \left \| Z_i - Z_j \right \|$.
        \item The relation between task $\tau_i$ and $\tau_j$ is determined by the distance between the representations $Z_i$ and $Z_j$ with a bandwidth $\sigma $, i.e., $m_{i,j}=\left \{ \left \| Z_i-Z_j \right \| <\sigma \right \}  $.
        \item The head $\hat{h}_i$ from the well-learned model $\mathcal{F}_\theta^{*}$ such that $\mathbb{E} \left[ (\hat{h}_i(g(x))-h_{i}(g(x)))^2 \right] =\mathcal{O}(\frac{\mathcal{R}(\mathcal{H})}{N_i^{tr}} ) $ where $\mathcal{R}(\mathcal{H})$ is the Rademacher complexity of the class $\mathcal{H}$.
    \end{itemize}
\end{assumption}

Next, we break down and explain each part of this assumption. All the conditions within this assumption are commonly used in the machine learning community \cite{mohri2018foundations}.

The first condition is about the uniform distribution of representations, i.e., \emph{$Z_i$ is assumed to be uniformly distributed on $\left [ 0,1 \right ]^k $}. This assumption asserts that for each task \(\tau_i\), the representation \(Z_i\) lies within the unit hypercube \([0, 1]^k\), where \(k\) is the dimensionality of the representation. This is one of the most commonly used assumptions in machine learning \cite{mohri2018foundations}. It suggests that the representations across tasks are spread evenly in this space. The uniform distribution assumption helps simplify the analysis of how the model generalizes across tasks. 

The second condition is about bounded distance between task representations, i.e., \emph{There exists a universal constant $C$ such that for all $i, j \in N_{tr}$, we have $\left \| h_{i} - h_{j} \right \|_{\infty }  \le  C \cdot  \left \| Z_i - Z_j \right \|$}. This assumption states that the distance between the task-specific heads \(h_i\) and \(h_j\) is bounded by a constant \(C\) times the distance between their corresponding representations \(Z_i\) and \(Z_j\). The \(\infty\)-norm denotes the maximum difference across each coordinate of the representations. It connects the geometry of the task representations (through \(Z_i\) and \(Z_j\)) with the behavior of the task-specific heads \(h_i\) and \(h_j\). If the representations are close, the corresponding task heads are also close, ensuring smooth transitions and generalization between tasks.
 
The third condition is about task similarity, i.e., \emph{The relation between task $\tau_i$ and $\tau_j$ is determined by the distance between the representations $Z_i$ and $Z_j$ with a bandwidth $\sigma $, i.e., $m_{i,j}=\left \{ \left \| Z_i-Z_j \right \| <\sigma \right \}  $}. It means that the relation between two tasks \(\tau_i\) and \(\tau_j\) is determined by the distance between their representations. Specifically, if the distance between \(Z_i\) and \(Z_j\) is smaller than a predefined threshold \(\sigma\), the tasks are considered similar. The variable \(m_{i,j}\) is a binary indicator that indicates whether tasks are similar. This assumption establishes a connection between the task representations and their perceived similarity, which is also a commonly used assumption.

The fourth condition is about head learning error and Rademacher complexity, i.e., \emph{The head $\hat{h}_i$ from the well-learned model $\mathcal{F}_\theta^{*}$ such that $\mathbb{E} \left[ (\hat{h}_i(g(x))-h_{i}(g(x)))^2 \right] =\mathcal{O}(\frac{\mathcal{R}(\mathcal{H})}{N_i^{tr}} ) $ where $\mathcal{R}(\mathcal{H})$ is the Rademacher complexity of the class $\mathcal{H}$}. This assumption states that the expected squared error between the learned head \(\hat{h}_i\) and the true head \(h_i\) is bounded by a term that scales with the Rademacher complexity \(\mathcal{R}(\mathcal{H})\) of the hypothesis class \(\mathcal{H}\) and inversely with the number of training examples \(N_i^{tr}\) for task \(\tau_i\). The Rademacher complexity captures the capacity of the model class to fit random noise, which is related to its ability to generalize. This assumption ties the learning error of the model to the complexity of the hypothesis class. It suggests that as the number of training examples increases, the model's learned head will get closer to the true task-specific head, and the error decreases. This is a typical assumption in generalization theory \cite{mohri2018foundations}.

\subsection{Proof of Theorem \ref{theorem:motivation}}

In the analyses, we consider a simple scenario involving two binary classification tasks, denoted as $\tau_i$ and $\tau_j$. That is, we set batchsize for training as 2. The label variables for these tasks are represented by $Y_i$ and $Y_j$, respectively, while $X_i$ and $X_j$ denote the sample variables for the two tasks. Given that these are binary classification tasks, $Y_i$ and $Y_j$ belong to the set of task labels $\{ \pm 1 \}$. It is worth noting that any multi-classification task can be decomposed into a combination of binary tasks (one against the other classes). In this proof, we focus on binary tasks to demonstrate the task confounder more simply and directly. Meanwhile, despite the two tasks are sampled from the same distribution, in this proof, we assume that these labels are drawn from two different probabilities, and the sampling probabilities of label values are balanced, i.e., $P(Y=1)=P(Y=-1)=0.5$. Our conclusions also hold for imbalanced distributions.

Given the set of causal factors for the entire world, $\mathrm{a}^w$, the training set represents a subset of the world with causal factors $\mathrm{a}^{tr} \subseteq \mathrm{A}^w$. Since $\mathrm{a}^{tr}$ is unknown, we model $\mathrm{a}^w$ using a Gaussian distribution, where the probability of a causal factor indicates its likelihood of belonging to $\mathrm{a}^{tr}$. For tasks $\tau_i$ and $\tau_j$, we consider two non-overlapping sets of factors, ${\rm a}^i$ and ${\rm a}^j$, representing knowledge in $N_z$ dimensions. These factors are assumed to be drawn from Gaussian distributions, i.e., ${\rm a}^i\sim \mathcal{N}(Y_i\cdot \mu_i,\sigma_i^2 I)$ and ${\rm a}^j\sim \mathcal{N}(Y_j\cdot \mu_j,\sigma_j^2 I)$. Here, $\mu_i,\mu_j\in \mathbb{R}^{N_z} $ denote the mean vectors, while $\sigma_i^2$ and $\sigma_j^2$ denote the covariance vectors.

In this analysis, we focus on the links of different task-specific model, which reflect the performance of meta-learning model $\mathcal{F}_{\theta}$ and decide whether to update further. For the sake of simplicity, we define $p$ to represent the varying correlations resulting from different task adaptations across different batches. Hence, we get:
\begin{equation}
\begin{array}{l}
    P(Y_i=Y_j)=p\\[8pt]
    P(Y_i\neq Y_j)=1-p
\end{array}
\end{equation}
When $p$ equals 0.5, it indicates that under this circumstance, the two tasks $\tau_i$ and $\tau_j$ are correlated within these environments. The objective of meta-learning adaptation is to obtain two linear models, $f_\theta^i: P(Y_i|{\rm a}^i,{\rm a}^j)$ and $f_\theta^j: P(Y_j|{\rm a}^i,{\rm a}^j)$ for $\tau_i$ and $\tau_j$.

Next, if the task-specific model promote each other, then the optimal classifier for each task has non-zero weights for non-causal factors, i.e., the task-specific factors of another task. When training $\mathcal{F}_{\theta}$ using two tasks, the optimal classifier for the target task will include causal features from the other task that are non-causal factors for the target task. To demonstrate this, we assume the use of a Bayesian classifier. Using task $\tau_i$ as an example, we can derive  the probability of $P(Y_i,{\rm A}^i,{\rm A}^j)$ with the optimal Bayesian classifier $P(Y_i|{\rm A}^i,{\rm A}^j)$ as follows:
\begin{equation}
\begin{array}{l}
     P(Y_i,{\rm a}^i,{\rm a}^j) = P(Y_i,{\rm a}^i)\cdot P({\rm a}^j|Y_i,{\rm a}^i)\\[8pt]
        = P(Y_i,{\rm a}^i)\cdot P({\rm a}^j|Y_i)\\[8pt]
        = P(Y_i,{\rm a}^i)\cdot\sum_{Y_j\in \left \{ -1,1 \right \} }P({\rm a}^j,Y_j|Y_i)\\[8pt]
        = P(Y_i)P({\rm a}^i|Y_i)\cdot\sum_{Y_j\in \left \{ -1,1 \right \} }P({\rm a}^j|Y_j)P(Y_j|Y_i)
\end{array}
\end{equation}
where the optimal Bayesian classifier $P(Y_i|{\rm A}^i,{\rm A}^j)$ is:
\begin{equation}\label{eq:P}
    P(Y_i|{\rm a}^i,{\rm a}^j) =\frac{P(Y_i,{\rm a}^i,{\rm a}^j)}{P({\rm a}^i,{\rm a}^j)} =\frac{P(Y_i,{\rm a}^i,{\rm a}^j)}{ {\textstyle \sum_{Y_i\in \left \{ -1,1 \right \}  }} P(Y_i,{\rm a}^i,{\rm a}^j)} 
\end{equation}
Assuming both ${\rm a}^i$ and ${\rm a}^j$ are drawn from Gaussian distributions, we have $P(Y_{i/j}, {\rm a}^i, {\rm a}^j) = \text{sigmoid}\left(\frac{\mu_i}{\sigma^2_i} {\rm a}^i + \frac{\mu_j}{\sigma^2_j} {\rm a}^j\right)$, where $\frac{\mu_i}{\sigma^2_i}$ and $\frac{\mu_j}{\sigma^2_j}$ are the regression vectors for the optimal Bayesian classifier. 

Then, instead of assuming a direct inclusion of both factors, we assume that the weights for \({\rm a}^j\) are modulated by the similarity \(\text{sim}(X_i, X_j)\).
To model this, we introduce a scaling factor based on the similarity between the tasks:
\[
\zeta = \text{sim}(X_i, X_j) \cdot \left( \frac{\mu_i}{\sigma_i^2} \cdot {\rm a}^i + \frac{\mu_j}{\sigma_j^2} \cdot {\rm a}^j \right)
\]
This scaling adjusts the impact of the task-specific factors from \(\tau_j\) on the classification of \(\tau_i\). As a result, the classifier's decision-making process for \(Y_i\) depends on both the task-specific factors from \(\tau_i\) and \(\tau_j\), weighted by their similarity. When the tasks are highly similar (\(\text{sim}(X_i, X_j)\) is large), the influence of \({\rm a}^j\) increases, leading to a stronger coupling between the tasks.

Combined with the assumptions that both ${\rm a}^i$ and ${\rm a}^j$ are drawn from Gaussian distributions, let $\zeta^{+}=\frac{\mu _i}{\sigma^2_i}{\rm A}^i+\frac{\mu _j}{\sigma^2_j}{\rm A}^j$ and $\zeta^{-}=\frac{\mu _i}{\sigma^2_i}{\rm A}^i-\frac{\mu _j}{\sigma^2_j}{\rm A}^j$. 
Thus, we first obtain:
\begin{equation}\label{eq:P_range}
\begin{array}{l}
     P(Y_i,{\rm a}^i,{\rm a}^j) = P(Y_i,{\rm a}^i)\cdot P({\rm a}^j|Y_i,{\rm a}^i)\\[8pt]
     = P(Y_i)P({\rm a}^i|Y_i)\cdot\sum_{Y_j\in \left \{ -1,1 \right \} }P({\rm a}^j|Y_j)P(Y_j|Y_i) \\[8pt]
     \propto e^{Y_i\cdot \frac{\mu _i}{\sigma^2_i}{\rm a}^i}(p e^{Y_i\cdot \frac{\mu _j}{\sigma^2_j}{\rm a}^j}+(1-p )e^{-Y_i\cdot \frac{\mu _j}{\sigma^2_j}{\rm a}^j}) \\[8pt]
     =p e^{Y_i\cdot (\frac{\mu _i}{\sigma^2_i}{\rm a}^i+\frac{\mu _j}{\sigma^2_j}{\rm a}^j)}+(1-p )e^{Y_i\cdot (\frac{\mu _i}{\sigma^2_i}{\rm a}^i-\frac{\mu _j}{\sigma^2_j}{\rm a}^j)}
\end{array}
\end{equation}
Then the Bayesian classifier $P(Y_i|{\rm A}^i,{\rm A}^j)$ becomes:
\begin{equation}
\begin{aligned}
    P(Y_i|{\rm a}^i,{\rm a}^j) =\frac{1}{1+\frac{p e^{Y_i\cdot \zeta ^{+}}+(1-p )e^{Y_i\cdot \zeta^- }}{p e^{-Y_i\cdot \zeta ^{+}}+(1-p )e^{-Y_i\cdot \zeta^- }} } 
\end{aligned}
\end{equation}
Next, when $p=0.5$, i.e., the correlation between $Y_i$ and $Y_j$ is equal to 0.5, we get:
\begin{equation}\label{eq:app_mo_1}
\begin{aligned}
    P(Y_i|{\rm a}^i,{\rm a}^j) =\frac{1}{1+e^{Y_i\cdot (\zeta^+ +\zeta^-)}}
    =\frac{1}{1+e^{2Y_i\cdot (\frac{\mu _i}{\sigma^2_i}{\rm a}^i)}}
\end{aligned}
\end{equation}
When $p\ne0.5$, i.e., the correlation between $Y_i$ and $Y_j$ is not equal to 0.5, we get:
\begin{equation}\label{eq:app_mo_2}
\begin{aligned}
    P(Y_i|{\rm a}^i,{\rm a}^j) =\frac{1}{1+e^{2Y_i\cdot \zeta ^+}} =\frac{1}{1+e^{2Y_i\cdot \zeta ^+}} 
\end{aligned}
\end{equation}
In both conditions, the optimal classifier for $\tau_i$ has non-zero weights for task-specific factors of $\tau_j$ with importance $\zeta $: (i) In Eq.\ref{eq:app_mo_1}, the optimal classifier for task $\tau_i$ only utilizes its factor ${\rm A}^i$ and assigns zero weights to the non-causal factor ${\rm a}^j$ which belongs to task $\tau_j$; (ii) In Eq.\ref{eq:app_mo_2}, the optimal classifier is both for the two factors ${\rm a}^i$ and ${\rm a}^j$. Thus, Theorem \ref{theorem:motivation} is certified.

\subsection{Proof of Theorem \ref{theorem:1}}

To establish the proof of Theorem \ref{theorem:1}, we initially define a function that serves as an intermediary, which can be expressed as:
\begin{equation}
    h^{in}_p = \frac{\sum_{i=1}^{N_{tr}} m_{ip} h_i} {\sum_{j=1}^{N_{tr}} m_{jp}}.
\end{equation}

We proceed to delineate an event, denoted as $\mathbf{\text{e}}_{N_{sh}} $, which is characterized by the condition $\sum_{i=1}^{N_{tr}} m_{ip} > 0$. Given our presupposition that:
\begin{equation}
    \scalebox{0.83}{$\mathbb{E}\left[(\mathcal{F}_{\theta}^*(x) -\mathcal{F}_{\theta}(x))^2\right]= \mathbb{E}\left[(h^*_i(g(x)) -h_i(g(x)))^2\right]= \mathcal{O}\left(\frac{\mathcal{R}(\mathcal{H})}{N^{tr}_i}\right)$},
\end{equation}
where $g$ denotes the feature extractor and $h$ denotes the classifier head for meta-learning. Here, also given the relationship $N^{tr}_i \gtrsim N_{sh}$ for every task $\tau_i$, it follows that during the occurrence of $\mathbf{\text{e}}_{N_{sh}} $, the following inequality holds:
\begin{equation}
\begin{split}
    &\mathbb{E}\left[(h^{in}_p(g(x)) - h^*_p(g(x)))^2\right]\\[8pt]
    &\le \frac{\sum_{i=1}^{N_{tr}} m_{ip} \cdot \mathbb{E} \left[(h^*_i(g(x)) - h_i(g(x)))^2\right]}{(\sum_{j=1}^{N_{tr}} m_{jp})^2} \\[8pt]
    &\le \frac{\max_i \mathbb{E}\left[(h_i^*(g(x)) - h_i(g(x)))^2\right]}{\sum_{j=1}^{N_{tr}} m_{jp}} \\[8pt]
    &= \mathcal{O}\left(\frac{\mathcal{R}(\mathcal{H})}{N_{sh} \sum_{j=1}^{N_{tr}} m_{jp}}\right).
\end{split}
\end{equation}

Furthermore, given that $\| h_i - h_j \|_\infty \le C \cdot \|Z_i - Z_\| \le C \cdot \sigma$ when $\|Z_i - Z_j\| \le \sigma$, we can assert that within the scenario $\mathbf{\text{e}}_{N_{sh}} $, the inequality $\left|h^{in}_k - h_k\right| \le C \cdot \sigma$ is valid. Conversely, for the complementary event $\mathbf{\text{e}}_{N_{sh}}^c $, the denominator is nullified by definition, rendering $h^{in}_k(g(x)) = 0$ and thus:
\begin{equation}
    \scalebox{0.83}{$\left|h^{in}_p(g(x)) - h_p(g(x))\right|^2 = \left(h_p\right)^2(g(x))
    \le (C \cdot \sigma)^2 + \left(h_p\right)^2(g(x)) \cdot \mathbf{1}_{\mathbf{\text{e}}_{N_{sh}}^c}$}.
\end{equation}
As a result, we derive that:
\begin{equation}
\begin{split}
     \mathbb{E}\left[(h^*_p - h_p)^2\right] \lesssim \mathbb{E}\left[\frac{\mathcal{R}(\mathcal{H})}{N_{sh} \sum_{j=1}^{N_{tr}} m_{jp}} \cdot \mathbf{1}_{\mathbf{\text{e}}_{N_{sh}} }\right] \\[8pt]     
     + \sigma^2 + \mathbb{E}\left[\left(h_p\right)^2(g(x)) \cdot \mathbf{1}_{\mathbf{\text{e}}_{N_{sh}} ^c}\right].
\end{split}
\end{equation}
For the initial term, let $S = \sum_{i=1}^{N_{tr}} \mathbf{1}\left\{\|Z_k - Z_i\| < \sigma\right\}$. Considering $Z^{un}$ are uniformly distributed over $[0,1]^p$, $S$ follows a binomial distribution $\mathcal{B}(N_{tr}, \varepsilon)$, where $\varepsilon = \mathbb{P}(\|Z - Z_k\| < \sigma)$. Utilizing the properties of the binomial distribution, we establish that:
\begin{equation}
    \mathbb{E}\left[\frac{\mathbf{1}\{S > 0\}}{S}\right] \lesssim \frac{1}{N_{tr} \varepsilon} \lesssim \frac{1}{N_{tr} \sigma^k}.
\end{equation}
Hence, the initial term is bounded by:
\begin{equation}
    \mathbb{E}\left[\frac{\mathcal{R}(\mathcal{H})}{N_{sh} \sum_{j=1}^{N_{tr}} m_{jp}} \cdot \mathbf{1}_{\mathbf{\text{e}}_{N_{sh}}}\right] \lesssim \frac{\mathcal{R}(\mathcal{H})}{N_{sh} N_{tr} \sigma^k},    
\end{equation}
The third term can be bounded in a similar fashion:
\begin{equation}
\begin{array}{l}
    \mathbb{E}\left[\left(h_p\right)^2(g(x)) \cdot \mathbf{1}_{\mathbf{\text{e}}_{N_{sh}}^c}\right] \\[8pt]
    \le \sup (h_p)^{2}(g(x)) \mathbb{E}[(1-q)^{N_{tr}}]\\[8pt]
    \lesssim \sup (h_p)^{2}(g(x))\frac{1}{N_{tr}q}\\[8pt]
    \lesssim \frac{1}{N_{tr} \sigma^k}.   
\end{array}
\end{equation}
By amalgamating all components, we arrive at:
\begin{equation}
    \mathbb{E}\left[(h^*_p - h_p)^2\right] \lesssim \sigma^2 + \frac{\mathcal{R}(\mathcal{H})/N_{sh}}{N_{tr} \sigma^k}.
\end{equation}
Given that $\ell$ is Lipschitz continuous with respect to its first parameter, the following inequality is obtained:
\begin{equation}
\begin{array}{l}
\mathbb{E}_{(x,y) \sim P_t} \left[\ell(\hat{f}_\theta^{(t)}(x), y)\right] - \mathbb{E}_{(x,y) \sim P_t} \left[\ell(f_\theta^{(t)}(x), y)\right] \\[8pt]
\le \mathbb{E}\left[|\hat{h}^{(t)} - h^{(t)}|\right] \\[8pt]
\le \sqrt{\mathbb{E}\left[(\hat{h}^{(t)} - h^{(t)})^2\right]}\\[8pt]
\lesssim \sigma + \sqrt{\frac{R(\mathcal{H})/N_{sh}}{N_{tr} \sigma^k}}.
\end{array}
\end{equation}
So far, we have completed the proof of Theorem \ref{theorem:1}.

\subsection{Proof of Theorem \ref{theorem:2}}

If we treat all training tasks as equally important, meaning $ m_{ip} = 1 $ for all $ \tau_i $ and $ \tau_p $, we can express the estimator $h_p$ as:
\begin{equation}
     h_p= \frac{\sum_{i=1}^{N_{tr}} m_{ip} h_p}{\sum_{j=1}^{N_{tr}} m_{jp}} = \frac{1}{N_{tr}} \sum_{i=1}^{N_{tr}} h_p.
\end{equation}
To show that this estimator performs worse than the optimal estimator $h^*_p$ in the minimax sense which is the classifier head of the trained model $\mathcal{F}_{\theta}^*$, we need to find an $h \in \mathcal{H}$ such that $ \mathcal{R}_h(h^{sum}(f(x))) = \Omega(1) $ even as $N_i^{tr}, N_{tr} \to \infty$. Here, we denote $h^{sum}$ as the average estimator of all tasks.

Consider the following setting: let $d \sim \mathcal{U}(0,1)$ be uniformly distributed on $(0,1)$ and let $g(x) \sim \mathcal{N}(0, 1)$ be normally distributed with mean 0 and variance 1. Define $h^d(g(x)) = d \cdot g(x)$. Under this setting, the average estimator $h^{sum}$ becomes $h^{sum} = \frac{1}{2} g(x)$ since the expectation of $d$ over a uniform distribution on $[0, 1]$ is $ \frac{1}{2} $. To compute the risk, we calculate:
\begin{equation}
    \scalebox{0.95}{$\mathbb{E}[(h^{sum}(g(x)) - h^d(g(x)))^2] = \mathbb{E}\left[\left(\frac{1}{2} g(x) - d \cdot g(x)\right)^2\right]$}.
\end{equation}
Simplifying inside the expectation, we get:
\begin{equation}
    \mathbb{E}\left[\left((\frac{1}{2} - d) g(x)\right)^2\right] = \mathbb{E}[(\frac{1}{2} - d)^2] \cdot \mathbb{E}[g(x)^2].
\end{equation}
Since $\mathbb{E}[g(x)^2] = 1$ (the variance of $g(x)$), we need to find $\mathbb{E}[(\frac{1}{2} - d)^2]$:
\begin{equation}
    \mathbb{E}\left[\left(\frac{1}{2} - d\right)^2\right] = \int_0^1 \left(\frac{1}{2} - d\right)^2 \, \mathrm{d}d.
\end{equation}
Evaluating this integral, we have:
\begin{equation}
\begin{array}{l}
    \mathbb{E}\left[\left(\frac{1}{2} - d\right)^2\right] \\[8pt]
    = \int_0^1 \left(\frac{1}{4} - d + d^2\right) \, \mathrm{d}d\\[8pt]
    = \left[\frac{1}{4}d - \frac{d^2}{2} + \frac{d^3}{3}\right]_0^1 \\[8pt]
    = \frac{1}{4} - \frac{1}{2} \cdot \frac{1}{2} + \frac{1}{3} = \frac{1}{12}.
\end{array}
\end{equation}
Therefore, we get:
\begin{equation}
    \mathbb{E}[(h^{sum}(g(x)) - h^d(g(x)))^2] = \frac{1}{12} = \Omega(1).
\end{equation}
Since the model $\mathcal{F}_{\theta} = h \circ g $ and the excess risk with task relation matrix $\mathcal{M}$ is denoted by $r(\mathcal{F}_{\theta}^*,\mathcal{M})=\sum_{(x,y)\in\mathcal{D}^{te}}\left [ \ell(\mathcal{F}_{\theta}^*(x),y;\mathcal{M} )-\ell(\mathcal{F}_{\theta}(x),y;\mathcal{M}) \right ]$, we have:
\begin{equation}
    \underset{\mathcal{F}_{\theta}^*}{\inf} \underset{h \in \mathcal{H}}{\sup} r(\mathcal{F}_{\theta}^*,\mathcal{M})-\underset{\mathcal{F}_{\theta}^*}{\inf} \underset{h \in \mathcal{H}}{\sup} r(\mathcal{F}_{\theta}^*,\check{\mathcal{M}}) <0.    
\end{equation}
This completes the proof.

\section{Practical Implementation of Meta-Learning}
\label{sec:O}

For optimization, the standard interpretation, i.e., understanding meta-learning from ``learning a good model initialization'', treats meta-learning as a second-order derivative process, while in practice, single-level updates are commonly used. Specifically, meta-learning models are typically updated via implicit gradients \cite{implicit1, implicit2, implicit3, implicit4, implicit5}, differentiable proxies \cite{differentiable1, differentiable2, differentiable3, differentiable4}, or single-layer approximations \cite{reptile, metaaug, nichol2018first} (\textbf{Appendix \ref{sec:O}}), which aggregate multi-tasks gradients into a single optimization step. 

For a simple but clear explanation, we set the parameters of the inner loop as $\phi$ and the parameters of the outer loop as $\theta$ according to the concept of ``learning to learn''. 
Then, the inner optimization problem is assumed to be $\phi^*(\theta) = \arg\min_{\phi} \mathcal{L}_{\text{inner}}(\theta, \phi)$, where $\theta$ is the outer parameter, $\phi$ is the parameter of inner loop, and $\mathcal{L}_{\text{inner}}$ is the loss function.

\paragraph{Implicit Gradient} The implicit gradient method calculates the outer gradient $\nabla_\theta \mathcal{L}_{\text{outer}}(\theta, \phi^*(\theta))$ by solving the following equation:
\begin{equation}
    \nabla_\theta \phi^*(\theta) = -\left[\nabla_{\phi}^2 \mathcal{L}_{\text{inner}}(\theta, \phi^*(\theta))\right]^{-1} \nabla_{\theta \phi}^2 \mathcal{L}_{\text{inner}}(\theta, \phi^*(\theta)),
\end{equation}
where $\nabla_{\phi}^2 \mathcal{L}_{\text{inner}}$ is the Hessian matrix with respect to $\phi$ for the inner loss function. Then, the outer gradient is computed using the chain rule:
\begin{equation}
    \scalebox{0.8}{$\nabla_\theta \mathcal{L}_{\text{outer}}(\theta, \phi^*(\theta)) = \nabla_{\phi} \mathcal{L}_{\text{outer}}(\theta, \phi^*(\theta)) \cdot \nabla_\theta \phi^*(\theta) + \nabla_{\theta} \mathcal{L}_{\text{outer}}(\theta, \phi^*(\theta))$}    
\end{equation}
This allows the outer parameter $\theta$ to be updated without explicitly solving the inner loop optimization.

\paragraph{Differentiable Proxies} In some applications, the inner optimization objective $\mathcal{L}_{\text{inner}}(\theta, \phi)$ may be difficult to compute or non-differentiable. To simplify this, a differentiable proxy function $\tilde{\mathcal{L}}_{\text{inner}}(\theta, \phi)$ can be used as a substitute:
\begin{equation}
\tilde{\mathcal{L}}_{\text{inner}}(\theta, \phi) \approx \mathcal{L}_{\text{inner}}(\theta, \phi).
\end{equation}
Then, the proxy function is used for inner loop optimization:
\begin{equation}
\phi^*(\theta) = \arg\min_{\phi} \tilde{\mathcal{L}}_{\text{inner}}(\theta, \phi).
\end{equation}
The outer loop optimization still targets $\mathcal{L}_{\text{outer}}(\theta, \phi^*(\theta))$.

\paragraph{Single-level Approximation} The bi-level optimization problem is simplified. The inner loop optimization becomes:
\begin{equation}
\phi^i = \phi - \alpha \nabla_{\phi} \mathcal{L}_{\text{inner}}(\theta, \phi),
\end{equation}
and the outer loop optimization is:
\begin{equation}
\min_\theta \sum_{i=1}^N \mathcal{L}_{\text{outer}}(\theta, \phi^i).
\end{equation}
In the single-level approximation, the update from the inner loop optimization is treated as a fixed value, and $\phi$ is no longer iteratively optimized. The outer loop directly uses the updated $\phi^i$ with:
\begin{equation}
\min_\theta \sum_{i=1}^N \mathcal{L}_{\text{outer}}(\theta, \phi - \alpha \nabla_{\phi} \mathcal{L}_{\text{inner}}(\theta, \phi)).
\end{equation}
In the optimization process of meta-learning model $\mathcal{F}_\theta$, the gradient information on all tasks is integrated into a single optimization step and directly used to update the global parameter $\theta$, which means that the model $\mathcal{F}_\theta$ is not gradually learned and optimized through internal and external loops, but directly adapted to multiple tasks through a single process. Therefore, the actual meta-learning model update method is more like a single-layer optimization process rather than true "learning to learn", i.e., it does not strictly follow the theoretical two-layer optimization framework.

\section{More Discussion}
\label{sec:discussion}

\subsection{Universality and Effectiveness of Task Relations}
TRLearner uses task relations and relation-aware consistency regularization to refine the meta-learning optimization. It assumes that similar tasks often share similar predictive functions, and thus enforces the outputs of task-specific models for similar tasks to be similar. Notably, although TRLearner computes task similarity to build a task relationship matrix, it does not require all tasks in a batch to be similar. Instead, TRLearner’s strength lies in leveraging inter-task relationships to highlight useful information—effectively filtering task information—and can work with any combination of tasks. Here, we discuss performance under both diverse and homogeneous task conditions from two angles: task construction in meta-learning and the sources of TRLearner’s effectiveness.

According to Section \ref{sec:3}, we assume all tasks—including the meta-training dataset \(\mathcal{D}_{tr}\) and the meta-test dataset \(\mathcal{D}_{te}\)—are drawn from the same fixed distribution \(p(\mathcal{T})\), with no class overlap between \(\mathcal{D}_{tr}\) and \(\mathcal{D}_{te}\). According to Baxter’s multi-task learning theorem \cite{baxter1997bayesian}, when a set of tasks sampled i.i.d. from the same distribution share certain structural commonalities (e.g., a common hypothesis space or representation structure), learning these tasks jointly can yield a lower-complexity representation space \cite{standley2020tasks}. This implies that the construction of meta-tasks itself provides theoretical support for TRLearner’s learning.
Furthermore, based on analyses by Maurer and Pentina \cite{maurer2016benefit}, if the meta-learner encounters multiple tasks from the same distribution during meta-training and obtains a ``low-complexity'' unified representation, then subsequent tasks drawn from the same distribution can achieve better generalization with fewer samples.
In addition, we build an adaptive sampler following \cite{wang2024towards} to obtain meta-data and extract task relationships. By selecting tasks under constraints of within-class compactness and between-class separability, the sampler makes it more challenging to extract structural commonalities among tasks. Under these conditions, leveraging task relations enables the model to better capture the underlying structural commonalities of the tasks. Thus, regardless of the initial task distribution, TRLearner can well calibrate the meta-learning optimization process and guide it to obtain effective representations.

Secondly, TRLearner’s effectiveness lies in its ability to filter and emphasize task information through inter-task relationships, rather than directly learning task-to-task similarities. Under limited data conditions, TRLearner leverages task relationships derived from meta-data to provide the model with additional insights, preventing it from over-focusing on task-specific features and thus maintaining a balance with over-parameterized networks. When data is abundant or tasks are diverse, these relationships guide the model to focus on shared, effective information across tasks. Grounded in causal invariance theory \cite{pearl2009causality}, such shared information also proves beneficial for downstream tasks. Consequently, even when tasks are largely unrelated, TRLearner can increase the weight of inter-task relationships to suppress task-specific factors (e.g., environmental features) and highlight shared factors (e.g., entity-related features) that enhance generalization to downstream tasks.
Moreover, both theoretical and empirical evidence from Section \ref{sec:5} and Section \ref{sec:7} demonstrate that TRLearner remains effective across various benchmark datasets and task distributions without relying on prior assumptions about the task distribution.

\subsection{How Task Relation Works From Task Information}

In meta-learning, the balance between task generality and task complementarity is crucial. Task generality refers to the task-shared knowledge between different tasks, which allows the model to learn general features and generalize to unseen tasks. For example, different classes may share similar visual features, e.g., edges, textures, or shapes. By identifying these task-shared features, the model can adapt to unseen tasks more quickly \cite{wilder2020learning}. In contrast, task complementarity refers to the relationship between different tasks, whereby learning this, the model can acquire more discriminative knowledge. This complementarity can help the model identify and utilize effective features to improve performance on specific tasks \cite{zheng2021knowledge}. For example, in multi-task learning, a model may learn classification and detection tasks at the same time. The classification task may help the model learn the general features of the object, while the detection task may emphasize the location and size of the object. This complementary knowledge learned by the model can improve the overall performance of the model on both tasks. 

Therefore, on the one hand, the model needs to be able to identify and utilize common knowledge across tasks to adapt to new tasks quickly; on the other hand, the model also needs to be able to learn task-specific knowledge to improve performance and accuracy on specific tasks. However, most existing methods focus on task generality and ignore task complementarity, which may cause the model to ignore important discriminative features and damage model performance.
In this study, one reason for how introducing task relations works is to use the power of task relations to force the model to learn task complementarity, which has been ignored in the past. Task relations cover the similarities or correlations between different tasks. To illustrate this concept, we take the drug response prediction task as an example: identifying each cell line is considered a separate task. If these cell lines show similar gene expression profiles or belong to the same cancer type, they are considered to be related, that is, there exist task relations. 

\subsection{More Discussion about Uniqueness of TRLearner}
\label{sec:app_comparison_field}

TRLearner is the first approach to leverage inter-task relationships to guide the optimization process in meta-learning from the perspective of learning lens, a consideration overlooked by previous work. While other fields have also explored introducing task-level information to improve model performance—such as cross-task validation and many-shot knowledge distillation—there are essential conceptual differences between these methods and TRLearner. Previous methods often focus on intra-task category relationships or use task performance as a validation tool. In contrast, TRLearner directly incorporates inter-task relationships into the optimization process, thus avoiding the pitfall of indiscriminately absorbing all information.

More specifically, cross-task validation \cite{anderson1972cross, kung2023active, wang2024meta, appel2021cross,kim2023cross} typically involves using the performance of auxiliary tasks during the meta-learning training phase to validate and regulate the main task’s learning. Essentially, it serves as an evaluation and monitoring mechanism during training rather than transmitting inter-task knowledge to enhance the model’s adaptability to new tasks.
Many-shot knowledge distillation \cite{sauer2022knowledge,li2021self,zhao2023mdcs,iscen2021class} mainly emphasizes aggregating knowledge from multiple sources (e.g., multiple teacher models or tasks with varying scales and perspectives) to provide rich informational inputs to a student model. However, it does not specifically emphasize selective filtering of task information or the utilization of inter-task structural relationships. As a result, the model may indiscriminately absorb all incoming information, increasing optimization difficulty and potentially introducing irrelevant features.
In contrast, TRLearner’s key innovation lies in its deep exploration and exploitation of inter-task correlations and commonalities, enabling the method to selectively and conditionally incorporate multi-task information. By leveraging an explicit inter-task relational structure, TRLearner can extract useful information from other tasks when data is scarce, thereby improving learning efficiency. Conversely, when data is abundant, it can filter out irrelevant factors and focus attention on the shared information most beneficial for the new task. Under varying data conditions, TRLearner consistently achieves stronger generalization and robustness. This feature not only emphasizes effective utilization of multi-task information but also ensures that optimization is guided by inter-task relationships, thus establishing a uniquely efficient paradigm for knowledge transfer and adaptation in meta-learning.

\section{Datasets}
\label{sec:app_C}
In this section, we elucidate the datasets encompassed within the four experimental scenarios.

\subsection{Regression}
We select the Regression problem with two datasets as our inaugural experimental scenario, i.e., Sinusoid and Harmonic datasets. The datasets here consist of data points generated by a variety of sinusoidal functions, with a minimal number of data points per class or pattern. Each data point comprises an input value $ x $ and its corresponding target output value $ y $. Typically, the input values for these data points fluctuate within a confined range, such as between 0 and $ 2\pi $.

In our experiment, we enhance the complexity of the originally straightforward problem by incorporating noise. Specifically, for Sinusoid regression, we adhere to the configuration proposed by \cite{jiang2022role,wang2023hacking}, where the data for each task is formulated as $ A\sin(\omega \cdot x) + b + \epsilon $, with $ A $ ranging from 0.1 to 5.0, $ \omega $ from 0.5 to 2.0, and $ b $ from 0 to $ 2\pi $. Subsequently, we introduce Gaussian observational noise with a mean of 0 and a variance of 0.3 for each data point derived from the target task. Similarly, the Harmonic dataset \citep{lacoste2018uncertainty} is a synthetic dataset sampled from the sum of two sine waves with different phases, amplitudes, and a frequency ratio of 2: $f(x) = a_1 \sin(\omega x + b_1) + a_2 \sin(2\omega x + b_2)$, where $y \sim \mathcal{N}(f(x), \sigma_y^2)$. Each task in the Harmonic dataset is sampled with $\omega \sim \mathcal{U}(5, 7)$, $(b_1, b_2) \sim \mathcal{U}(0, 2\pi)^2$, and $(a_1, a_2) \sim \mathcal{N}(0, 1)^2$. This process finalizes the construction of the dataset for this scenario.

\subsection{Image Classification}
For our second scenario, image classification, we select four benchmark datasets with two experimental settings, including standard few-shot learning (SFSL) with miniImagenet \cite{miniImagenet,lin2023multi,zhang2024metadiff} and Omniglot \cite{Omniglot}, and cross-domain few-shot learning (CFSL) with CUB \cite{cub} and Places \cite{places}. We now provide an overview of the four datasets in this scenario.
\begin{itemize}
    \item miniImagenet consists of 50,000 training images and 10,000 testing images, evenly spread across 100 categories. The first 80 of these categories are designated for training, while the final 20 are reserved for testing, with the latter never encountered during the training phase. All images are sourced from Imagenet.
    \item Omniglot is designed to foster the development of learning algorithms that mimic human learning processes. It encompasses 1,623 unique handwritten characters from 50 distinct alphabets, each drawn by 20 different individuals through Amazon's Mechanical Turk. Each character image is paired with stroke data sequences $ [x, y, t] $ and temporal coordinates (t) in milliseconds.
    \item CUB is extensively utilized for tasks involving the fine-grained differentiation of visual categories. It encompasses a collection of 11,788 photographs, categorized into 200 distinct bird species, with 5,994 images designated for training purposes and 5,794 for testing. The dataset provides comprehensive annotations for each photograph, including a single subcategory label, the precise locations of 15 parts, 312 binary attributes, and a single bounding box. We split all the data into 100/50/50 classes for meta-training/validation/testing. 
    \item Places is a comprehensive image collection designed for the task of scene recognition, a critical area within the field of computer vision. It boasts an extensive library of over 2.5 million images, meticulously categorized into 205 unique scene categories. Each image is meticulously curated to represent a wide array of natural and man-made environments, providing a rich tapestry of visual data for training and evaluating machine learning models. We split them into 103/51/51 classes for meta-training/validation/testing.
\end{itemize}
Note that the models in the SFSL setting are trained and tested on their evaluation datasets, while the models in the CFSL setting are trained on the miniImagenet dataset and then tested on the CUB or Places datasets.

\subsection{Drug Activity Prediction}
In our third scenario, concerning drug activity prediction, we align with the data partitioning delineated in \cite{martin2019all,jiang2022role}. We extract 4276 tasks from the ChEMBL database \cite{gaulton2012chembl} to constitute our baseline dataset, which is preprocessed in accordance with the guidelines set forth by \cite{martin2019all}.

ChEMBL is a comprehensive database utilized extensively in chemical biology and drug research, housing a wealth of biological activity and chemical information. It contains over 1.9 million compounds, more than 2 million bioactivity assay results, and thousands of biological targets, all meticulously structured. This includes the structural data of drug compounds, bioactivity assay outcomes, and descriptions of drug targets. Following the approach in \cite{martin2019all}, we segregate the training compounds in the support set from the testing compounds in the query set, with the meta-training, meta-validation, and meta-testing task distribution being 4100, 76, and 100, respectively.

\subsection{Pose Prediction}
For our final scenario, we select the Pascal 3D dataset \cite{xiang2014beyond} as our benchmark and process it accordingly. We randomly select 50 objects for meta-training and an additional 15 objects for meta-testing.

The Pascal 3D dataset is composed of outdoor images, featuring 12 classes of rigid objects chosen from the PASCAL VOC 2012 dataset, each annotated with pose information such as azimuth, elevation, and distance to the camera. The dataset also includes pose-annotated images for these 12 categories from the ImageNet dataset. For the pose prediction task, we preprocess it to include 50 categories for meta-training and 15 for meta-testing, with each category comprising 100 grayscale images, each measuring $ 128 \times 128 $ pixels.

\section{Baselines}
\label{sec:app_D}
In this paper, we focus on the generalization of meta-learning and select four optimization-based meta-learning methods as the backbone for evaluating the performance of TRLearner, i.e., MAML \cite{maml}, MetaSGD \cite{Meta-sgd}, ANIL \cite{anil}, and T-NEt \cite{tnet}. Meanwhile, we also select multiple baselines for comparison, including regularizers which handle meta-learning generalization, i.e., Meta-Aug \cite{metaaug}, MetaMix \cite{yao2021improving}, Dropout-Bins \cite{jiang2022role} and MetaCRL \cite{wang2023hacking}, and the SOTA methods which are newly proposed for generalization, i.e., Meta-Trans \cite{bengio2019meta}, MR-MAML \cite{mrmaml}, iMOL \cite{imol}, OOD-MAML \cite{oodmaml}, and RotoGBML \cite{rotogbml}. Here, we briefly introduce all the methods used in our experiments.

\textbf{MAML (Model-Agnostic Meta-Learning)} is a widely used meta-learning algorithm that seeks to find a model initialization capable of being fine-tuned to new tasks with a few gradient steps. It focuses on learning an initialization that facilitates rapid adaptation.

\textbf{MetaSGD} is a meta-learning algorithm that adapts the learning rate during the meta-training process. It focuses on optimizing the learning rate, potentially improving the model's ability to generalize across tasks.

\textbf{ANIL} aims to reduce the number of inner loop iterations during meta-learning. It optimizes the meta-learner by minimizing the reliance on costly inner loop optimization steps, aiming for more efficient training.

\textbf{T-NET (Task-Agnostic Network)} learns a shared representation across tasks. It aims to develop a task-agnostic feature extractor that captures common patterns in different tasks, thereby improving generalization.

\textbf{ProtoNet (Prototypical Networks)} learns to map input data into an embedding space and represent each class by a “prototype,” which is the mean vector of its support samples in that space. During prediction, a new sample is also embedded and then classified by finding the class prototype closest to it in terms of distance. 

\textbf{Meta-Aug} is built by using data augmentation in the meta-learning process, with the goal of generating more diverse training samples to improve the generalization ability of the model. This includes common data augmentation techniques such as random cropping, rotation, and scaling.

\textbf{MetaMix} aimed at enhancing generalization in meta-learning tasks. It employs techniques to improve the model's capability to handle variations and adapt to new tasks more effectively.

\textbf{Dropout-Bins} utilizes dropout techniques to improve generalization in meta-learning. These techniques enhance model robustness and help mitigate overfitting.

\textbf{MetaCRL} is based on causal inference and explores the task confounder problem existing in meta-learning to eliminate confusion, improving the generalization and transferbility of meta-learning.

\textbf{Meta-Trans} combines transfer learning and meta-learning to fine-tune the pre-trained model to adapt to new tasks. The model is adjusted based on existing knowledge to improve the generalization ability on new tasks.

\textbf{MR-MAML} addresses the bias introduced by task overlap by designing a meta-regularization objective using information theory that prioritizes data-driven adaptation. This leads the meta-learner to decide what must be learned from the task training data and what should be inferred from the task test inputs.

\textbf{iMOL} is proposed for continuously adaptive out-of-distribution (CAOOD) detection, whose goal is to develop an OOD detection model that can dynamically and quickly adapt to emerging distributions and insufficient ID samples during deployment. It is worth noting that in order to adapt iMOL to the tasks of regression, classification, etc. in this paper, we rewrote the loss function of the method.

\textbf{OOD-MAML} is a meta-learning method for out-of-distribution data. It improves the generalization ability of the model by learning tasks on different distributions, especially when facing new distributions.

\textbf{RotoGBML} homogenizes the gradients of OOD tasks, thereby capturing common knowledge from different distributions to improve generalization. RotoGBML uses reweighted vectors to dynamically balance different magnitudes to a common scale, and uses rotation matrices to rotate conflicting directions to be close to each other.

These methods and backbones are critical components of the experimental setup and are used to construct a comprehensive empirical analysis in this paper.

\end{document}